%% file: root.tex
\documentclass[letterpaper, 10 pt, conference]{ieeeconf}  

\IEEEoverridecommandlockouts                              

\usepackage{amsthm}
\usepackage{amsmath}
\usepackage{graphicx}
\usepackage{xcolor}

\usepackage{amssymb}
\usepackage{hyperref}
\usepackage{makecell}
\usepackage{caption}
\usepackage{subcaption}

\usepackage{algorithm}
\usepackage{algorithmic}
\title{\LARGE \bf
On the Stability Analysis of Open Federated Learning Systems}

\author{Youbang Sun$^{1}$~~~ Heshan Fernando$^{2}$~~~ Tianyi Chen$^{2}$~~~ Shahin Shahrampour$^{1}$
\thanks{The work of Y. Sun and S. Shahrampour was supported by NSF ECCS-2136206 Award. The work of H. Fernando and T. Chen was supported by National Science Foundation CAREER Award 2047177.}
\thanks{$^{1}$Youbang Sun and Shahin Shahrampour are with the Department of Mechanical and Industrial Engineering, Northeastern University, Boston, Massachusetts, 02115, USA
        \{\tt\small sun.youb,  s.shahrampour\}@northeastern.edu}%
\thanks{$^{2}$Heshan Fernando and Tianyi Chen are with Department of Electrical, Computer, and Systems Engineering, Rensselaer Polytechnic Institute,        Troy, New York, 12180, USA
        \{\tt\small fernah,chent18\}@rpi.edu}%
}

\begin{document}

\input{header.tex}

\maketitle
\thispagestyle{empty}
\pagestyle{empty}
\linespread{0.99}

\begin{abstract}
    We consider the open federated learning (FL) systems, where clients may join and/or leave the system during the FL process.  Given the variability of the number of present clients, convergence to a fixed model cannot be guaranteed in open systems. Instead, we resort to a new performance metric that we term the stability of open FL systems, which quantifies the magnitude of the learned model in open systems. Under the assumption that local clients' functions are strongly convex and smooth, we theoretically quantify the radius of stability for two FL algorithms, namely local SGD and local Adam. We observe that this radius relies on several key parameters, including the function condition number as well as the variance of the stochastic gradient. Our theoretical results are further verified by numerical simulations on  synthetic data.
\end{abstract}

\section{Introduction}
Federated learning (FL) \cite{mcmahan2017communication} is a machine learning setup where a group of clients work cooperatively to learn a statistical model. The learning process is coordinated by a central server which facilitates the exchange of model updates. FL algorithms enjoy the benefits of model sharing among clients while preserving data privacy, and they also reduce the number of communications without making too much sacrifice on the performance \cite{woodworth2020local}. In a canonical FL algorithm, the central server broadcasts the initial model to all clients, and then, each client performs several steps of local updates before sending the model to the server. Once receiving models from a subset of clients, the server performs a global aggregation, typically in the form of average. The aggregated model will then be broadcasted again to all clients, and the algorithm continues.

FL emerges as a new paradigm to address the following four challenges in distributed machine learning \cite{mcmahan2017communication, li2020federated}. The first concern is \textit{communication efficiency}, where FL algorithms often reduce the communication frequency in distributed architectures \cite{konevcny2016federated}. The second problem is that the local objective functions are different across clients, which is a consequence of having different data-sets (and optimal solutions) in respective clients. This challenge is termed as \textit{statistical heterogeneity} \cite{li2020federatedoptimization}. The third challenge is \textit{systems heterogeneity}, which accounts for the devices (rather than data) varying from one client to another, including differences in processing speeds, communication bandwidth, and reliability \cite{wang2020tackling}. Lastly, \textit{privacy} is another concern in FL \cite{bhowmick2018protection}, which can be partially mitigated by preventing from directly transmitting local data to the central server.

\begin{figure}[t]
  \centering
  \includegraphics[width=\linewidth]{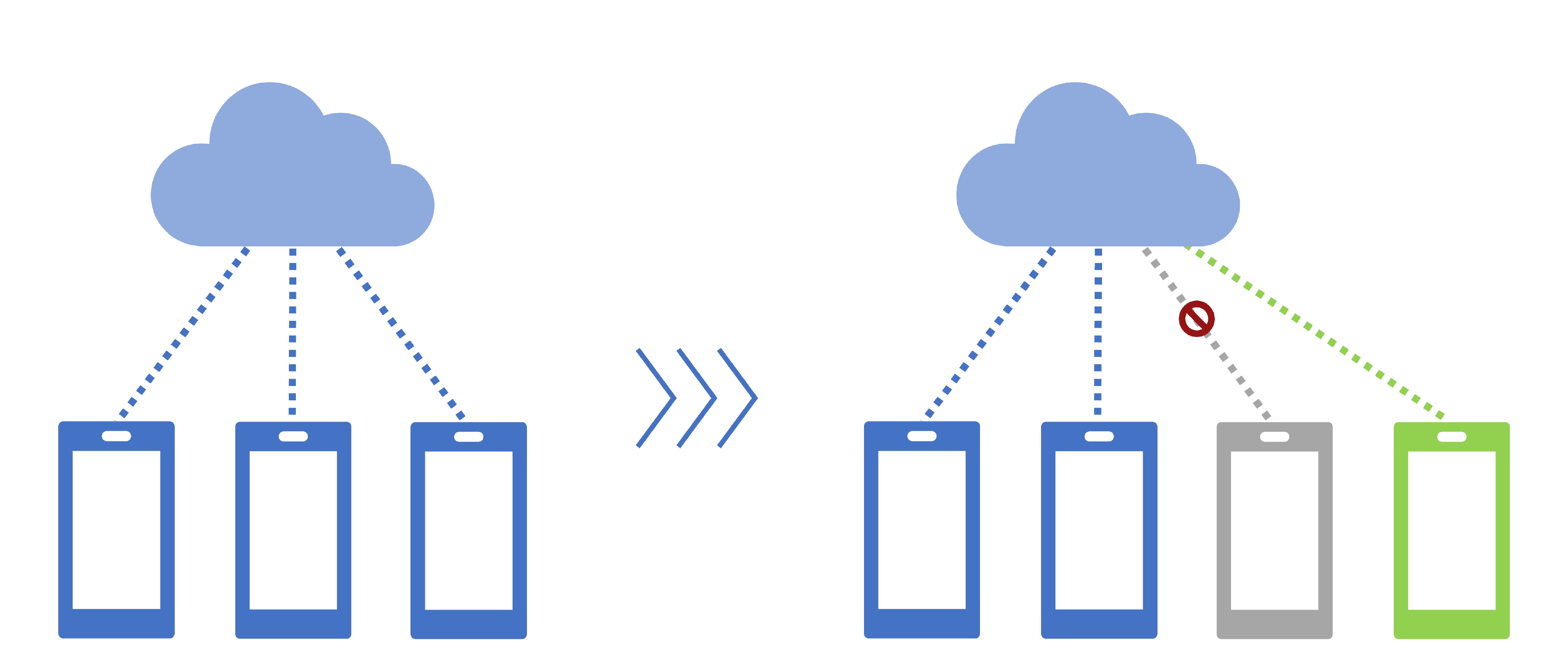}
  \caption{An illustration of the open FL system}
  \label{fig:open-system}
\end{figure}

In this work, we consider a new FL setting that we term the open FL system, where clients may join and/or leave the system during the learning process. More generally, the local clients' functions may change over time. Open FL systems are common in real-life scenarios. For example, a node failure in a large distributed system is almost inevitable, which translates to clients leaving the open FL system. In social platforms, users constantly join and leave. Also, it is possible that a present user might not be willing to share its model at a certain time. Therefore, addressing open systems is critical for FL architectures. 
There are recent works such as \cite{sahu2018convergence,reisizadeh2020fedpaq, karimireddy2020scaffold} that consider the client {\it drop-out}. However, these works mostly emphasize the partial cooperation of the clients, and the clients are still considered to be within the system. We view an open FL system through a different lens, where at certain times (when the system is still operating), new clients may join and old clients may leave the system; see an illustration in Fig. \ref{fig:open-system}.

In open FL systems, due to the possible change in local objective functions for various clients, the global objective function also varies over time. As a result, analyzing the convergence to an exact statistical model (i.e., fixed optimization solution) is not possible for this type of systems. Additionally, when a client joins or leaves the system, the network size (i.e., number of clients) increases or decreases, respectively, which poses another challenge. 
In order to address this issue, we focus on the \textit{stability} rather than the convergence of the open FL system. The concept of stability is precisely defined in Section \ref{setup_stability}, which quantifies the magnitude of the learned model in open systems.

Our contributions are as follows
\begin{itemize}
  \item \textbf{[Stability as a new FL metric]} We provide a formal definition of stability for open FL systems.
Then, we provide analysis on the global stability of the open FL system under the assumption that the local optimization methods are stable. 

  \item \textbf{[Stability of two FL algorithms]}  We focus on two most common optimization methods used for FL (local SGD and local Adam) and provide theoretical guarantee on their stability, respectively. 
  
\item \textbf{[Empirical verification of stability]} Experiments on both synthetic data as well as real world datasets are conducted, and our empirical results further verify our theoretical findings.

\end{itemize}

\subsection{Related Work}
We provide the literature review on several aspects of FL and distributed optimization. 

\noindent\textbf{Federated learning.}
Federated averaging (FedAvg) proposed by \cite{mcmahan2017communication} was the first algorithm that introduced the federated learning setup. Since then, many variants have been proposed for a variety of motivations. FedAvg can be simplified to LocalSGD proposed by \cite{zinkevich2010parallelized}, which has been thoroughly analyzed in recent works (\cite{stich2018local, stich2019error, khaled2020tighter, woodworth2020local}). Recently \cite{mitra2021achieving} established linear convergence while considering client heterogeneity and sparse gradients.
\cite{li2020federatedoptimization} considered a regularized objective function in order to deal with statistical heterogeneity. \cite{karimireddy2020scaffold} introduced SCAFFOLD, where the difference in clients are monitored to reduce the variance. \cite{yu2019linear} added momentum to the local clients and obtained better convergence results with the cost of extra communication. \cite{xie2019local, wang2021local} introduced adaptivity in local clients. A different approach to add adaptivity and acceleration to FedAvg was proposed by \cite{reddi2020adaptive,wang2019slowmo}, where the adaptivity is introduced in the server instead of clients. Apart from adaptive optimization methods, the communication can also be considered as adaptive, as shown in \cite{chen2021cada}. We refer to \cite{kairouz2021advances} for a more general overview of the literature.

\noindent\textbf{Open multi-agent networks.} In this literature, clients are often referred to as {\it agents}. The behavior of agents joining and leaving the system has long been studied by the community of distributed computing. \cite{article} proposed FIRE and \cite{carrascosa2009service} proposed THOMAS, both addressing this mechanism in computations. Due to the recent resurgence of distributed optimization motivated by machine learning, a new line of work focusing on the dynamics of the system has been proposed. \cite{8263752} provided analysis for open systems, and \cite{hendrickx2020stability} provided further analysis for distributed gradient descent under more general network assumptions. The open system setup has also been discussed for algorithms such as online optimization \cite{hsieh2022multi} and dual averaging \cite{hsieh2021optimization}. To the best of our knowledge, however, open FL systems have not been addressed. 

\noindent\textbf{Dynamical system stability.}
The stability of a dynamical system is an important topic in control and optimization. A very well-known notion of stability is the Lyapunov stability \cite{lyapunov1992general}. \cite{kozin1969survey} introduced the notion of Lyapunov stability in the $m$-th mean for stochastic systems, which is akin to our definition of stability in Section \ref{setup_stability}. Similar stability notions have also been discussed recently under different setups. \cite{hosoe2019equivalent} and \cite{hosoe2021second} considered the second-moment stability of discrete-time linear systems. The Markov jump system has also been discussed in \cite{costa2013stochastic}. However, all of these works touch upon the stability of the system dynamics rather than the stability of a FL architecture.

Another well-known line of work includes \cite{hardt2016train}. These works consider the stability of optimization methods in the sense of \cite{bousquet2002stability}, which is closely related to the algorithm generalization. These works are not directly related to the notion of stability in this paper, and we only provide this remark to note the difference. 

\section{Problem Formulation}

{\bf Notation:} We denote by $[n]$ the set $\{1,2,3,\ldots,n\}$ for any integer $n$. We use $\langle x, y \rangle$ to denote the standard inner product between $x$ and $y$, and $\norm{x} = \sqrt{\langle x, x \rangle}$ is the Euclidean norm of vector $x$.
We use $\col\{v_1, \ldots, v_n\}$ to denote the vector that stacks all vectors $v_i$ for $i\in [n]$.
{We use $\mathbf{1}_d$ to denote a $d$-dimensional vector consists of ones.}
We use $\diag \{a_1, \ldots, a_n\}$ to represent an $n\times n$ diagonal matrix that has the scalar $a_i$ in its $i$-th diagonal element. 
We denote the $k$-th element of a sequence $\{x^{(1)},x^{(2)},x^{(3)},... \}$ by $x^{(k)}$. We write $\mathbf{B}(x, r) = \{y: \norm{y - x} \leq r\}$ to denote a ball of radius $r$ centered at $x$. The $i$-th element of a vector $v$ is denoted as $[v]_i$. 

\subsection{Federated Learning Setup}\label{FL_setup}
In the canonical FL setup, we have a set of $N$ clients and one central server that together perform a distributed optimization. The ultimate goal for the entire FL architecture is to minimize the following global objective function
\begin{equation}
  \label{globalfunction}
  \underset{x \in \R^d }{\text{minimize}} \:\:\:\:\:\:F(x) =\sum_{i=1}^N f_i(x)=\sum_{i=1}^N \E[f_i(x, \xi_i)],
\end{equation}
which outputs a global statistical model learned by the entire network. However, consistent with privacy-preserving aspect of FL, each client $i\in [N]$ is associated with the local objective function $f_i(x):\R^d \rightarrow \mathbb{R}$ that encapsulates its own data $\xi_i$. Therefore, clients perform local updates and every once in a while communicate with the central server, which performs averaging among local models. We classify the iterations into two separate categories, namely, the communication iterations and the local update iterations, where during the communication iterations, the clients do not perform any local update.




\subsection{Open Federated Learning Architecture}
In this work, we are concerned with {\it open} FL systems, where clients may join or leave the system, or more generally, the local functions for clients may slightly change over time. A common real-life example is a social platform where users become online/offline frequently with a user base that evolves over time. We are inspired to consider this approach to FL architectures following the work of \cite{8263752} on open multi-agent systems. We note that when all local agents in the FL system perform gradient descent and communicate every round, our setup reduces to the open system distributed gradient descent formulation, which is addressed in \cite{hendrickx2020stability}.

In order to adapt FL algorithms to the open system setting, we need to make some adjustments to the FedAvg algorithm. If a client joins the system in between two communications, it will not be initialized since the server might not be available for broadcasting. When the next communication happens, the FedAvg algorithm does not consider the newly joined client as an {\it eligible} client, and this new client will only start its local update after receiving broadcasting from the server in the coming communication round. For the ease of analysis, we assume that the availability of agents (i.e., joining/leaving the system) is modeled with i.i.d. random variables, and there is at least one eligible client available for every communication. However, this i.i.d. assumption is not necessary.

\subsection{Local Optimization Methods}

Recent works have studied many forms of FL; one variation of the original FedAvg algorithm is to change the local optimizer from gradient descent to other optimization methods. Our main focus is to study open FL systems, and to that end, we consider two standard local optimization methods, namely stochastic gradient descent (SGD) and Adam. SGD is the workhorse of training for many machine learning models, commonly embedded into FL architecture for large-scale data \cite{mcmahan2017communication}. 
In SGD, the local clients perform the following update
\begin{equation}
    x_i^{(k+1)} = x_i^{(k)} - \eta \nabla f_i(x_i^{(k)}, \xi_i^{(k)}),
\end{equation}
where $k$ is the iteration index, and $i\in[N]$ is the client index, respectively. On the other hand, Adam can exhibit a superior performance due to its adaptive nature without significantly increasing time/space costs {\cite{kingma2014adam}}. In particular, for the Adam algorithm, at the $k$-th local iteration all local clients perform the following update
\begin{equation}
    \begin{aligned}
        h_i^{(k+1)} &= \beta_1 h_i^{(k)} + (1-\beta_1) \nabla f_i(x_i^{(k)}, \xi_i^{(k)})\\
        v_i^{(k+1)} &= \beta_2 \hat{v}_i^{(k)} + (1-\beta_2) ( \nabla f_i(x_i^{(k)}, \xi_i^{(k)}))^2\\
        x_i^{(k+1)} &= x_i^{(k)} - \frac{\eta}{(\epsilon I + \hat{V}_i^{(k+1)})^{1/2}}  h_i^{(k+1)},
    \end{aligned}
\end{equation}
where $v_i^{(k)} \in \R^{d}$ is a vector that estimates the second moment of the stochastic gradient, and $\hat{V}_i^{(k)} \in \R^{d \times d}$ is a diagonal matrix, such that $\hat{V}_i^{(k+1)}:=\diag\{\hat{v}_i^{(k+1)}\}$ and $\hat{v}_i^{(k+1)} := \max\{v_i^{(k+1)}, \hat{v}_i^{(k)} \}$.

\subsection{Stability in Open FL Systems}\label{setup_stability}
Given that in open FL architectures, clients may join and leave the system, exact convergence analysis is not possible. Here, we present the notion of stability in open FL systems with respect to the second-moment of optimization iterates. 
 
\begin{definition}\label{def:stability}
For an iterative stochastic optimization algorithm $\mathcal{A}$, which produces a sequence of iterates $\{x^{(k)}\}$, {we say that $\mathcal{A}$ is stable with respect to the second-moment} if there exist a finite constant $R$ and an iteration index $k_0$, such that 
\begin{center}
when $\norm{x^{(k)}}^2 \leq R^2$,\\ then
$\ex{\norm{x^{(k+1)}}^2 \big| x^{(k)}}\leq R^2$ for all $k \geq k_0$.      
\end{center}
\end{definition}

\subsection{Technical Assumptions}
In this paper, we impose the following assumptions on the local objective function.

\begin{assumption}\label{local_assumption}
For any client $i\in [N]$ in the network, we assume that the local objective function $f_i:\R^d \to\R$ is $\mu$-strongly convex and $L$-Lipschitz smooth.
\end{assumption}

The smoothness assumption is standard in FL systems. The strong convexity assumption is not typical, but we require that to analyze the dynamics of optimization iterates $x^{(k)}$. Without strong convexity, one cannot analyze the distance between $x^{(k)}$ and $x^\star$ given that the optimal solution might not be unique. Additionally, having both strong convexity and Lipschitz smoothness allows us to define the condition number $\kappa = \frac{L}{\mu} \geq 1$, which is crucial in the analysis of similar works \cite{hendrickx2020stability}. Note that the strongly convex part of Assumption \ref{local_assumption} implies that the global objective function $F(x)$ is also strongly convex, and hence, there exists a unique finite minimizer $x^\star$.

\begin{assumption}\label{optimal_ball_assum}
For any client $i\in [N]$ in the system, we denote the optimal solution of the local function $f_i(x)$ by $x_i^\star$, and we assume that $x_i^\star \in \mathbf{B}(0,r)$ with $f_i(x_i^\star) = 0$.
\end{assumption}
Since all local functions satisfy Assumption \ref{local_assumption}, Assumption \ref{optimal_ball_assum} holds without loss of generality by simply re-centering the functions. This assumption only simplifies our analysis.


 \begin{algorithm}[t]
\caption{Federated Learning in Open Systems}\label{algorithm:alg main text}
\begin{algorithmic}[1]
\STATE \textbf{client initialize:} Local optimizer with respective hyper-parameters, e.g. learning rate $\eta$.
\STATE \textbf{global initialize:} Initial model  $x^{(0)}$, broadcast model to  available clients, and mark all  clients eligible.
\STATE \textbf{for} $k = 1,2,\dots,K$, \textbf{do}:
\STATE \hspace{+0.3cm}\textbf{if} the iteration is local update iteration:
\STATE \hspace{+0.5cm}Eligible clients perform a local update.
\STATE \hspace{+0.3cm}\textbf{else}: 
\STATE \hspace{+0.5cm}Server randomly selects a subset of clients from eligible clients and averages local parameters.
\STATE \hspace{+0.5cm}Broadcast parameters to all clients.
\STATE \hspace{+0.5cm}Mark all clients eligible.
\STATE \hspace{+0.3cm}\textbf{end if}
\STATE \hspace{+0.3cm}Current clients randomly leave the system and become ineligible.
\STATE \hspace{+0.3cm}New clients randomly enter the system; initialize local optimizer and set client state to ineligible; wait for broadcasted model parameters.
\STATE \textbf{end for}
\end{algorithmic}
\end{algorithm}

\section{Main Results}\label{sec:main}

\subsection{System-wide Stability}
FL systems generally consist of two components. First is the local update (e.g., SGD or Adam), 
and second is the federated averaging, wherein the central server aggregates the information from a selection of local clients to construct an improved global model. In regards to the stability of open FL systems, our first result is to show that when the local update satisfies stability in the sense of Definition \ref{def:stability}, the stability over the entire open FL system can be claimed consequently.

\begin{proposition}\label{prop:stability}
Given Assumptions \ref{local_assumption}-\ref{optimal_ball_assum}, for a local optimizer $\mathcal{A}^{loc}$ defined in a Federated Learning system $\mathcal{A}^{Fed}$, if $\mathcal{A}^{loc}$ satisfies the stability in the sense of Definition \ref{def:stability}, then the entire system $\mathcal{A}^{Fed}$ is also stable in the sense that
\begin{center}
 when $\norm{x^{(k)}_i}^2 \leq R^2$ for all $i \in [N]$,\\ then
$\mathbb{E}[\norm{x^{(k+1)}_i}^2|x^{(k)}_i]\leq R^2$ for all $i \in [N]$, $k \geq k_0$.
\end{center}

\end{proposition}
\begin{proof}

(i) If the current iteration $k$ is a local update iteration, each of the clients performs a local update; therefore, stability follows from Definition \ref{def:stability}.  


(ii) If the current iteration $k$ is a communication iteration, consider the averaging setup discussed in Section \ref{FL_setup}. The system takes an average over a random subset of clients. Since the selection of clients is independent of client number, we can write the averaged result as $\Bar{x}^{(k)} = \frac{\sum_{i = 1}^N \delta_i x_i^{(k)}}{\sum_{i = 1}^N \delta_i}$, where $\delta_1,...,\delta_N$ are non-negative i.i.d. random variables. 
We note that for the vanilla FedAvg algorithm, $\delta_i$ follows a Bernoulli distribution though $\delta_i$ can also follow other distributions, which corresponds to the central server taking a  weighted average over clients. Then, after averaging and broadcasting to all agents, we have
\begin{equation}
    \begin{aligned}
        \mathbb{E}\left[ \norm{{x}^{(k+1)}_i}^2\right] &=   \mathbb{E}\left[ \left\| \frac{\sum_{i = 1}^N \delta_i x_i^{(k)}}{\sum_{i = 1}^N \delta_i}\right\|^2\right] \leq R^2.
    \end{aligned}
\end{equation}
Therefore, the system has stability on both local update iterations and communication iterations.
\end{proof}
Although the analysis of stability in Proposition \ref{prop:stability} was for system variable $x_i$ (i.e., client $i$), we note that this also provides us with insights about objective error. From Assumption \ref{optimal_ball_assum} we know that since the optimal solution $x_i^\star \in \mathbf{B}(0,r)$, when $x_i^{(k)} \in \mathbf{B}(0,R)$, the distance $\norm{x_i^{(k)} - x_i^\star}$ is upper bounded. Therefore, the optimization error is also upper bounded due to Lipschitz smoothness from Assumption \ref{local_assumption}.








\subsection{Open FL System Stability - SGD}
We now consider the stability of FL systems under the SGD algorithm, which is the most common algorithm used in FL. We impose the following assumption on the stochastic gradient of the objective function.

\begin{assumption}\label{sgd_grad_assumption}
    The stochastic gradient is unbiased with bounded variance, i.e. 
    $$\mathbb{E}\left[\nabla f_i(x_i^{(k)}, \xi_i^{(k)})\right] = \nabla f_i(x_i^{(k)})$$
    $$\mathbb{E}\left[\norm{\nabla f_i(x_i^{(k)}, \xi_i^{(k)}) - \nabla f_i(x_i^{(k)})}^2\right]\leq \sigma^2.$$
\end{assumption}
The assumption above is quite standard and essential for the analysis in many previous works on SGD under various settings, including
\cite{mcmahan2017communication, reisizadeh2020fedpaq,spiridonoff2020local} and more. In the following theorem, we quantify the radius of stability for open FL systems using SGD as the local optimizer. We remark that since our analysis holds for any available client, we present the result with a generic function $f(x)$, instead of a client-specific function $f_i(x)$.







\begin{theorem}\label{thm_sgd}
 For any function $f(x)$ that satisfies the Assumption \ref{local_assumption} with the optimal solution $x^\star \in \Bb(0,r)$, if the SGD algorithm runs with a learning rate $\eta \in (0, \frac{1}{L}]$, under Assumption \ref{sgd_grad_assumption} there exists a radius 
    $$R=r+\sqrt{{\max (3, \kappa)}}\sqrt{2r^2+\frac{{\sigma}^2}{{L}^2}},$$
for the FL system to have stability with respect to the second moment defined in Definition \ref{def:stability}.
\end{theorem}
The complete proof of this theorem is given in \cite{https://doi.org/10.48550/arxiv.2209.12307}. We can see that the radius depends on $\kappa$ and $\sigma$: if the stochastic gradient variance and/or the condition number increase, the radius also increases, which intuitively makes sense. For a special case where $\sigma = 0$, the FL algorithm will perform GD instead of SGD, and the stability results can be considered as a special case from \cite{hendrickx2020stability}, in a federated setup (i.e., centralized).

\subsection{Open FL System Stability - Adam}
We now quantify the radius of stability for open FL systems using Adam as the local optimizer. Similar to SGD, we first impose an assumption on the stochastic gradient used in the Adam algorithm, which is also commonly used in previous related works such as \cite{wang2021local}.

\begin{assumption}\label{adam_grad_assumption}
The stochastic gradient is unbiased and bounded, i.e. 
    $$\mathbb{E}\left[\nabla f_i(x_i^{(k)}, \xi_i^{(k)})\right] = \nabla f_i(x_i^{(k)})$$
    $$\norm{\nabla f_i(x_i^{(k)}, \xi_i^{(k)})}\leq \sigma.$$
\end{assumption}
Again, since our analysis holds for any available client, we present the result with a generic function $f(x)$, instead of a client-specific function $f_i(x)$.
\begin{theorem}\label{thm_adam}
    For any function ${f}(x)$ that satisfies Assumption \ref{local_assumption} with the optimal solution $x^\star \in \Bb(0,r)$, if the Adam algorithm runs with learning rate $\eta \in (0, \frac{1}{L}]$, under Assumption \ref{adam_grad_assumption} there exists a finite radius 
    \begin{equation}
    \begin{aligned}
         R = C_5 r + \sqrt{\frac{1+3\kappa C_1}{(1-\kappa C_1)^2}r^2 + \frac{{2}(C_2 + C_3 +  C_1 C_4)}{\mu - {L}C_1}},
    \end{aligned}
\end{equation}
    with constants $C_1 - C_5$ for the FL system to have stability with respect to the second moment defined in Definition \ref{def:stability}. 

\end{theorem}
We refer to \cite{https://doi.org/10.48550/arxiv.2209.12307} for explicit forms of constants $C_1-C_5$ and the proof of this theorem. We remark that $\sigma$ should be small enough for the radius to exist. More discussion can be found in \cite{https://doi.org/10.48550/arxiv.2209.12307} (Section \ref{sec:thm}).
\section{Numerical Experiments}
In order to verify our analysis, we conduct simulations using logistic regression on a synthetic dataset.  Data are uniformly distributed across a group of dynamic clients during all tests.  We implement local Adam and local SGD in the open FL setting, where clients perform local iterate updates independently, which are averaged periodically at a central server over all currently available clients. We specify the details of our simulation setting below.

\textbf{Open FL system setup.} We simulate a synthetic open FL system as follows. The open FL system consists of $N$ clients in total, each containing $m$ data. Initially, the system has $N_0$ clients that communicate with server. Each of the clients currently communicating with the server has a probability of leaving $p$ before the next communication round, and similarly there is a probability $p$ of a new client joining the system. Once a client leaves, it never joins the current system again. At the $k$-th communication round, the server communicates with $N_k$ clients in the system. The newly joined clients will upload their local iterates to the server only after receiving the global iterate from the server in the previous communication iteration. For simplicity we allow only one client to leave and one client to join the system, at any given global iteration. We assume $N_k>0$ for all $k=0, 1, 2, ...$, that is, the the system will consist of at least one client at all times.

 We plot the norm of the global iterate at each communication round under different choice of regularization $\lambda$ of local loss function, probability $p$ of clients joining/leaving the system, and in the synthetic setting the standard deviation of the data $\sigma$. We investigate the stability of the global iterate with changing $\lambda$, since $\lambda$ determines the condition number $\kappa$ of the loss function, which has a significant effect on $R$.

\begin{figure}[t]
\begin{subfigure}{0.235\textwidth}
  \centering
  \includegraphics[width=1\linewidth]{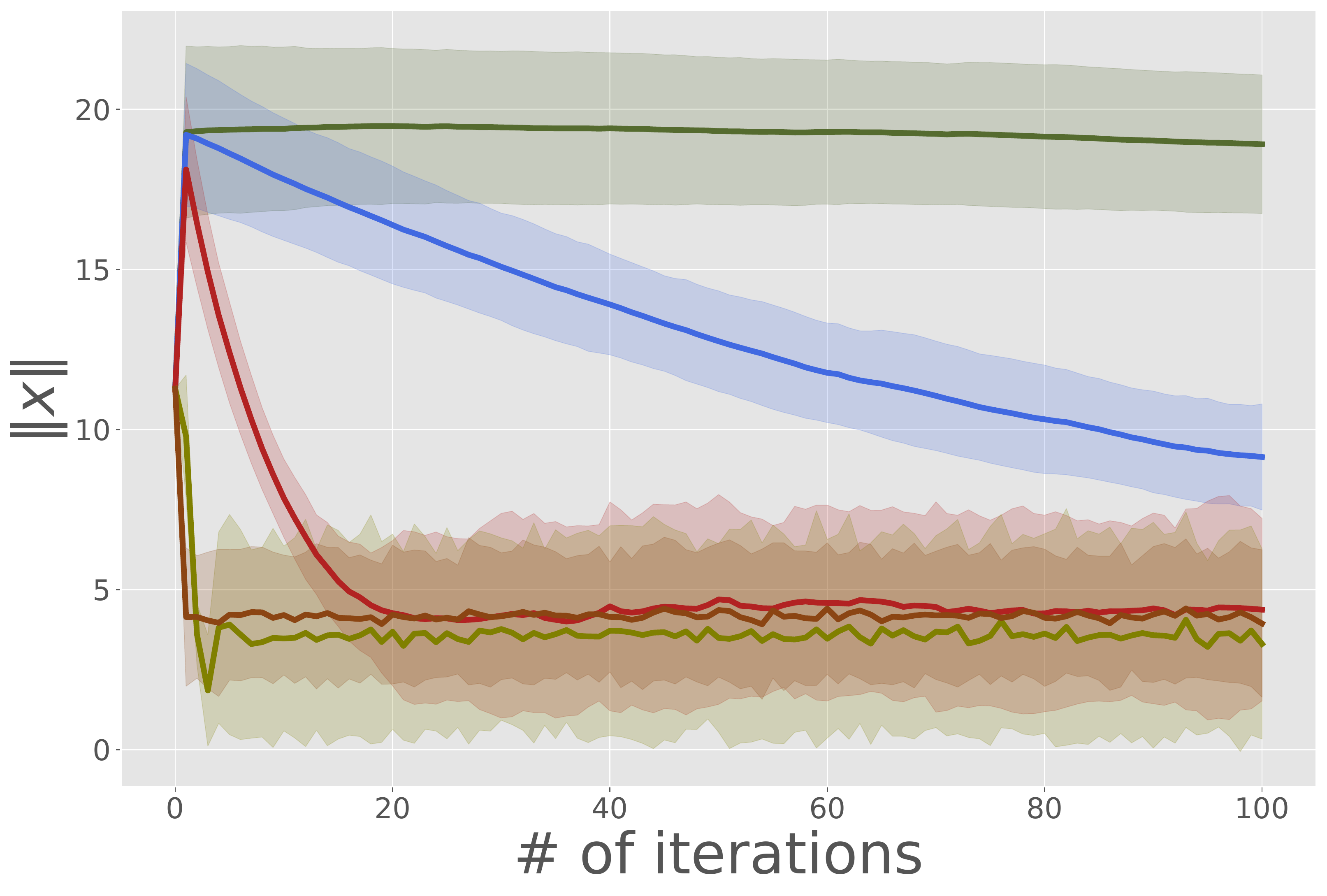}  
  \caption{local SGD}
\end{subfigure}
\hfill
\begin{subfigure}{0.235\textwidth}
  \centering
  \includegraphics[width=1\linewidth]{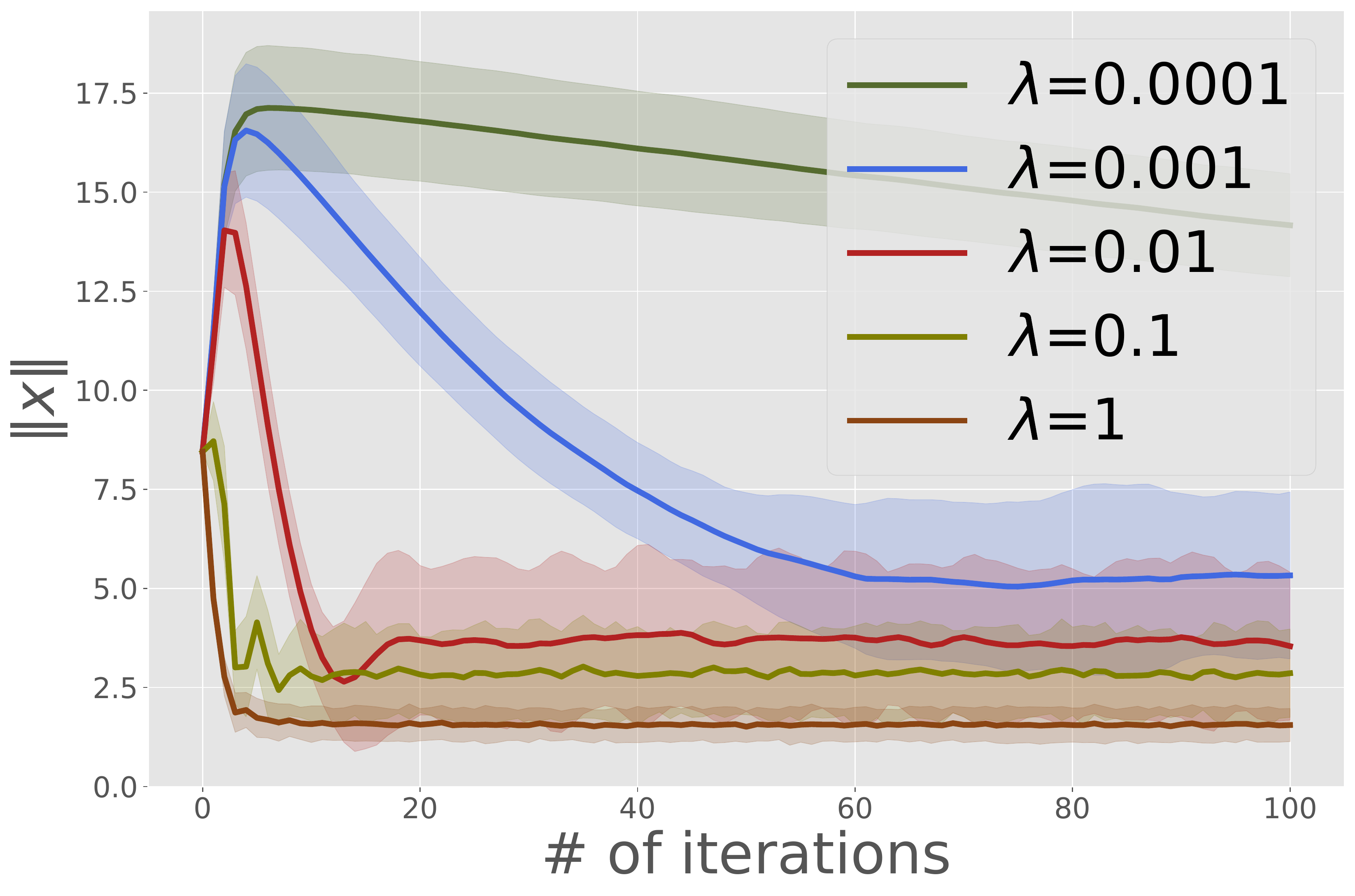}
  \caption{local Adam}
\end{subfigure}
\caption{Comparison of iterate norms for different $\lambda$.} 
\label{fig:synth-lambda-itnorms}
\end{figure}

\begin{figure}[t]
\begin{subfigure}{0.235\textwidth}
  \centering
  \includegraphics[width=1\linewidth]{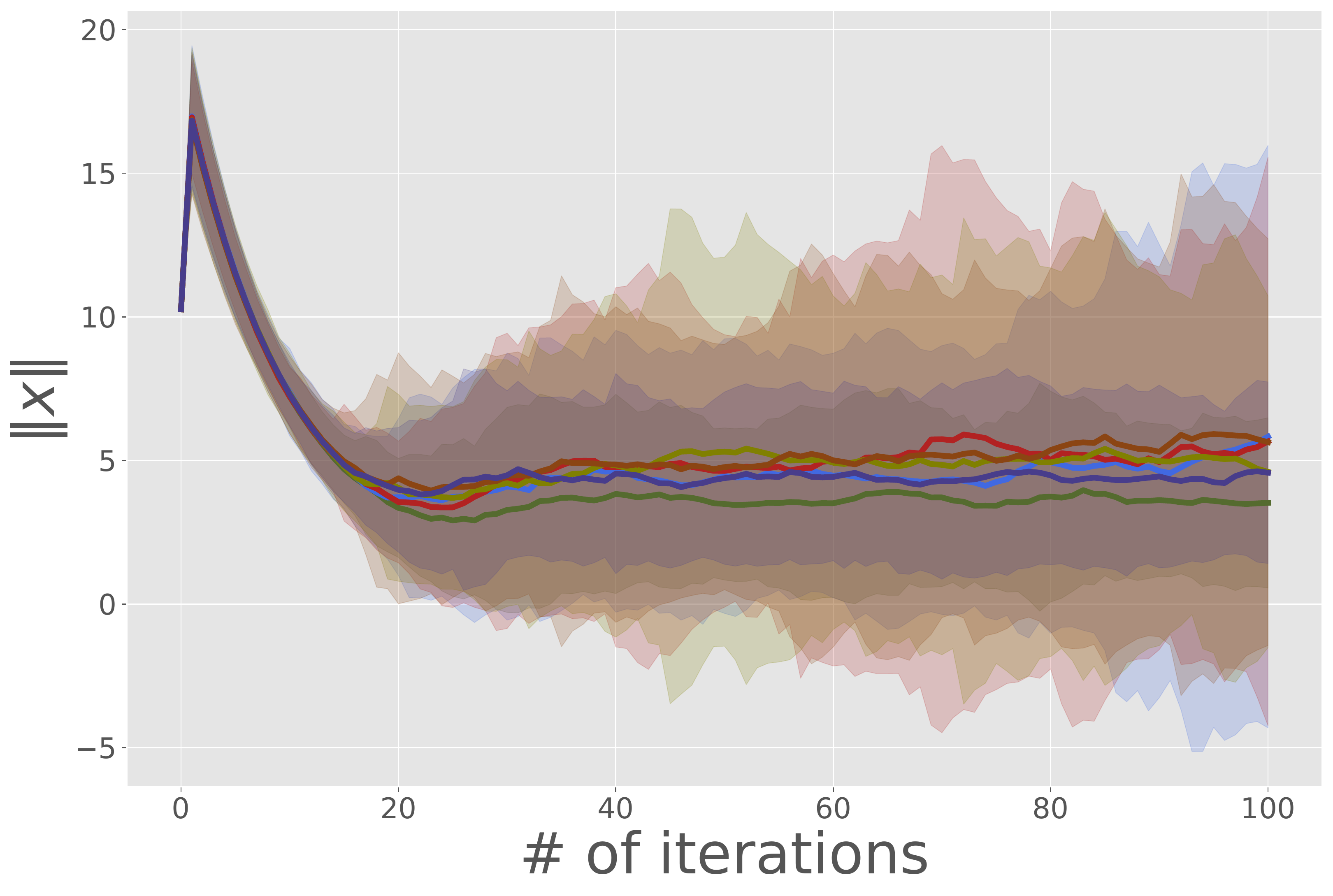}  
  \caption{local SGD}
\end{subfigure}
\hfill
\begin{subfigure}{0.235\textwidth}
  \centering
  \includegraphics[width=1\linewidth]{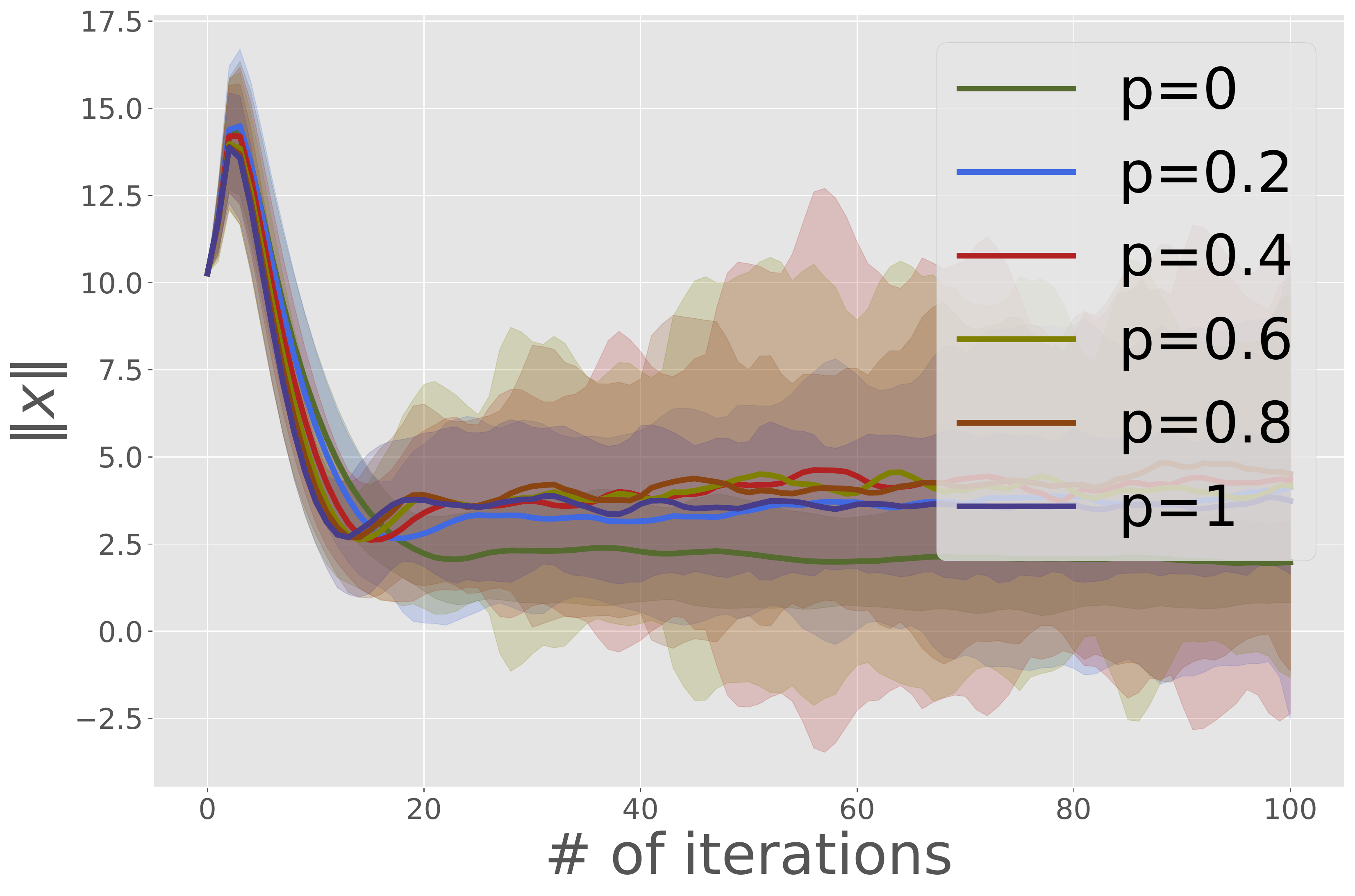}
  \caption{local Adam}
\end{subfigure}
\caption{Comparison of iterate norms for different $p$.} 
\label{fig:synth-p-itnorms}
\end{figure}

\begin{figure}[t]
\begin{subfigure}{0.235\textwidth}
  \centering
  \includegraphics[width=1\linewidth]{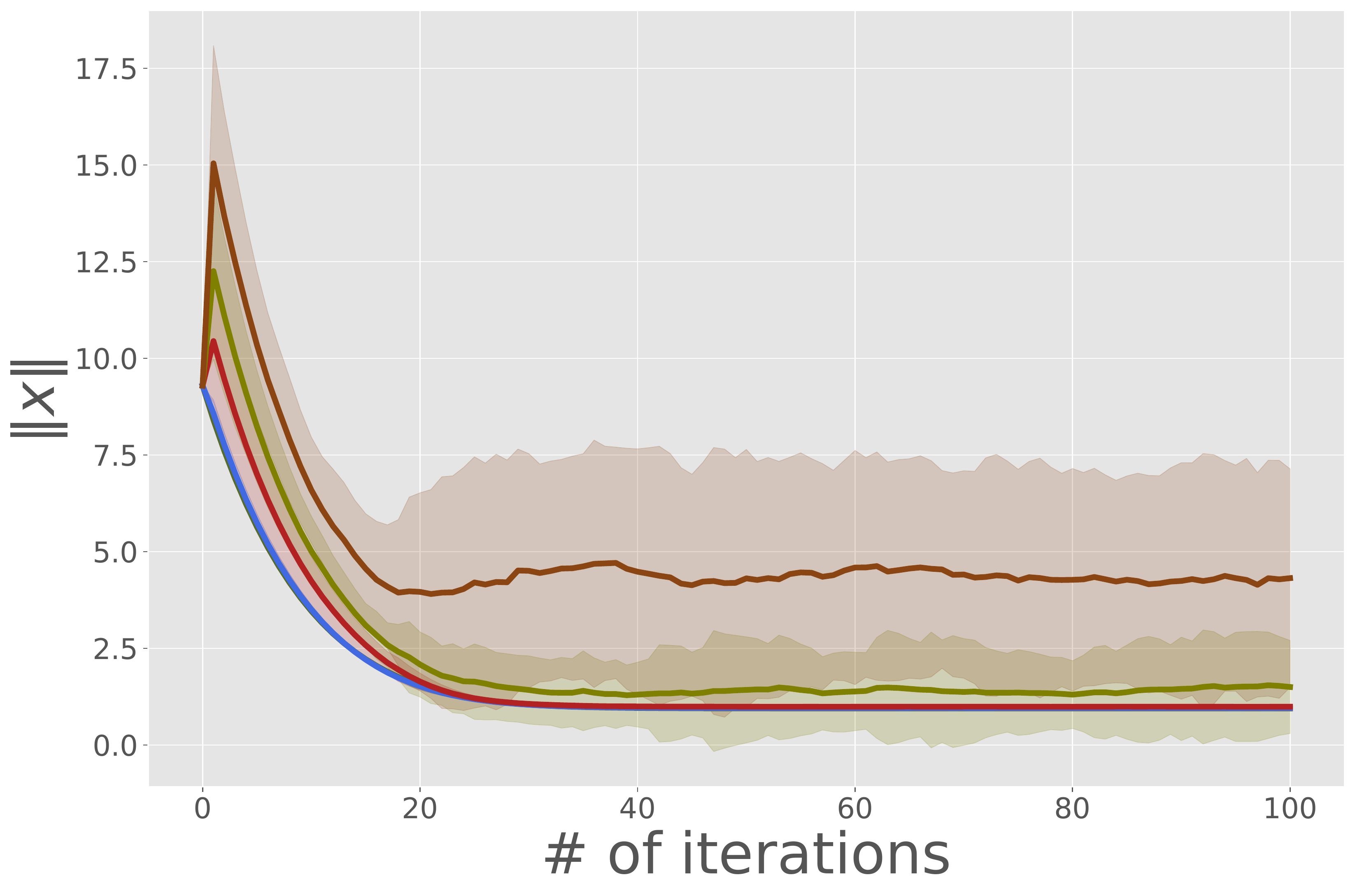}  
  \caption{local SGD}
\end{subfigure}
\hfill
\begin{subfigure}{0.235\textwidth}
  \centering
  \includegraphics[width=1\linewidth]{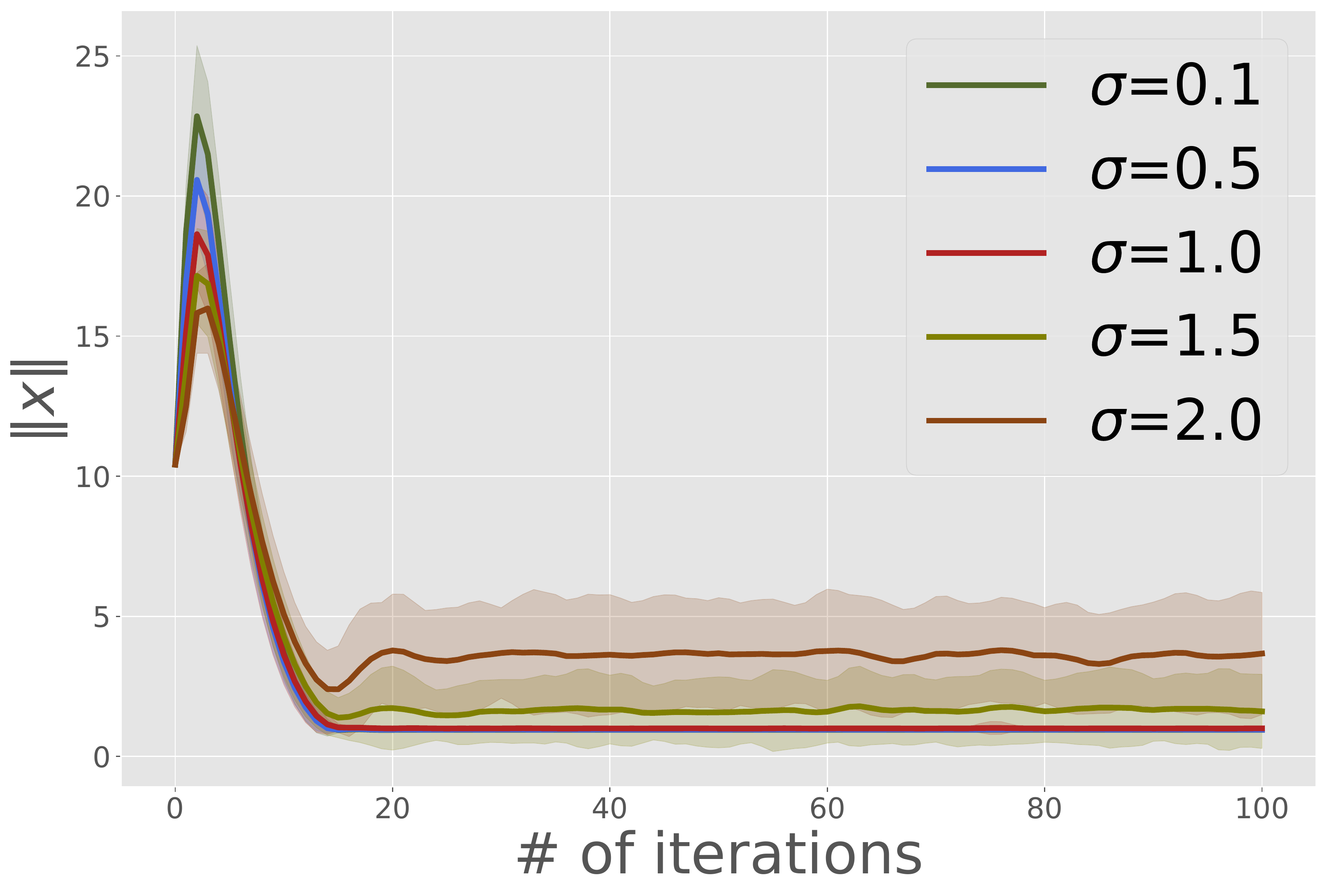}
  \caption{local Adam}
\end{subfigure}
\caption{Comparison of iterate norms for different $\sigma$.} 
\label{fig:synth-sigma-itnorms}
\end{figure}
\subsection{Synthetic setting}\label{sec:synth}

In the synthetic data setting, the clients will have data of dimension $d=100$ corresponding two label classes $y =-1$ and $y=1$. The two label classes corresponds to two Gaussian distributions constructed as follows: the mean vector for each label class is sampled as $\mu = 2{\rm Bern}(0.5)-1$ element-wise, and the standard deviation of the data (element-wise) is $\sigma$. Each client will have $m=100$ data.

We use the logistic loss as the local loss function at each client, given by
$$f_i(x)=\frac{\lambda}{2}\norm{x}^2+\frac{1}{m}\sum_{j=1}^m \log(1+\exp(-y_{i,j} x^\top \xi_{i,j})),$$
where $x$ is the local iterate, $\lambda$ is the regularization parameter, $\{(\xi_{i, j}, y_{i, j})\}_{j=1}^{m}$ is the local dataset.

For the synthetic data setting, we use $N_0=10$ and $N=1000$. We run 100 Monte-Carlo simulations for each choice of $p$, $\lambda$ and $\sigma$, and plot the average results. The local optimization parameters for local SGD are set as $\eta=1$, and for local Adam as $\beta_1=0.9$, $\beta_2=0.999$, $\epsilon=10^{-3}$ and $\eta=0.1$. Fig. \ref{fig:synth-lambda-itnorms} shows the norm of the global iterate for different $\lambda$, where the remaining system parameters are set to $p=1$ and $\sigma=2$. Here we can observe that in both local Adam and local SGD cases, the norm of the global iterate converge to a stable value faster when the regularization coefficient is larger. Fig. \ref{fig:synth-p-itnorms} shows the norm of the global iterate for different $p$, where the remaining system parameters are set to $\lambda=0.01$ and $\sigma=2$. Here, it can be observed that when the system is highly dynamic, i.e. when $p$ is closer to 1, the expectation of the stable value of the iterate is larger. The smallest global iterate norm corresponds to $p=0$, when the system is static.

Fig. \ref{fig:synth-sigma-itnorms} shows the global iterate norm for different $p$, where the remaining system parameters are set to $p=1$ and $\sigma=2$. It can be seen that when the variance of the data is high, the global iterate norm is larger. 

We further verify our results with extensive experiments in various settings including real-world data, where details of these experiments are provided in \cite{https://doi.org/10.48550/arxiv.2209.12307}. From the results for synthetic and real world data settings, the global iterate of the algorithms local SGD and local Adam maintains stability in systems with dynamic clients. The empirical results show how the norm of the global iterate changes with dynamic nature of the system and also with local loss function parameters such as $\lambda$ and $\sigma$. The experiment results align with the relationship between stability radius and function parameters given in Section \ref{sec:main}.






\section{Conclusions}

We introduced a novel formulation for FL, named open FL systems, where clients are able to join and/or leave the system during execution. We provided a formal definition for stability in open FL systems. We then provided theoretical analysis for stability for two commonly used optimization methods. The analytical results of this paper was further validated by numerical experiments on synthetic data.

\bibliographystyle{IEEEtran}

\bibliography{ref}

\newpage


 \input{supp_content}

\end{document}


\input{header.tex}

%

%

\onecolumn
\aistatstitle{Supplementary Material}



\section{Omitted Proofs}
We provide the details of proof for theorems and related lemmas, which was previously omitted in the paper.
\subsection{Proof of Theorem 2}
We first introduce the following lemma to replace the arbitrary learning rate $\eta$ with a fixed learning rate $\beta = \frac{1}{L}$.

\begin{lemma}\label{lemma_eta}
    If for variables $x, \Delta x$, there exists some finite scalar $R$ that satisfies $\norm{x} \leq R$ and $\norm{x - \frac{1}{L} \Delta x} \leq R$, then $\norm{x - \eta \Delta x} \leq R$ is true for any $\eta \in (0, \frac{1}{L}]$.
\end{lemma}
\begin{proof}
Since ${x} \in \mathbf{B}(0, R)$, and ${x - \frac{1}{L} \Delta x} \in \mathbf{B}(0, R)$, and that $\mathbf{B}(0, R)$ is a convex set, ${x - \eta \Delta x} $ which is a linear combination of the former two will also be contained in the same convex set.
        
\end{proof}

We can replace the learning rate in SGD with the maximum possible learning rate $\frac{1}{L}$ using Lemma \ref{lemma_eta}. Next we present our proof for Theorem 2.
\begin{proof}
We first assume that the condition number $\kappa  = \frac{L}{\mu} \geq 3$, note that when function has better a condition number (i.e. $\kappa < 3$), it can be easily treated and we consider its condition number as equal to $3$, since we can consider a smaller number $\mu' < \mu$, then a $\mu$-strongly convex function is always $\mu'$-strongly convex.

We consider the $k$-th iteration. From the strong convexity and smoothness assumptions of the objective function, we can write the following inequality,
\begin{equation}
    \begin{aligned}
        &\mathbb{E}\left[ \langle x^{(k)} - x^\star,  \nabla {f}(x^{(k)}; \xi^{(k)}) \rangle\right] = \langle x^{(k)} - x^\star, \nabla {f}(x^{(k)})\rangle\\
    &\geq \frac{{L}^{-1}||\nabla {f}(x^{(k)})||^2}{1+{\kappa}^{-1}} + \frac{\mu||x^{(k)}-x^\star||^2}{1+{\kappa}^{-1}}
    \end{aligned}
\end{equation}

Adding $\mathbb{E}\left[ \langle \nabla {f}(x^{(k)}; \xi^{(k)}) , x^\star \rangle  \right]$ to both sides, using Cauchy-Schwartz inequality and the assumption that $\norm{x^\star}\leq r$, we obtain
\begin{equation}
    \begin{aligned}
        &\mathbb{E}\left[  \langle x^{(k)},  \nabla {f}(x^{(k)}; \xi^{(k)}) \rangle \right]\\
        &\geq \frac{{L}^{-1}||\nabla {f}(x^{(k)})||^2}{1+{\kappa}^{-1}} + \frac{\mu||x^{(k)}-x^\star||^2}{1+{\kappa}^{-1}}-r \norm{\nabla {f}(x^{(k)})}
    \label{prod}
    \end{aligned}
\end{equation}

We focus on the case $\eta=\frac{1}{{L}}$. By the definition of SGD,
\begin{equation}
    \begin{aligned}
            &\mathbb{E}\left[\norm{x^{(k+1)}}^2\right] = \mathbb{E}\left[\norm{x^{(k)} - \frac{1}{{L}}\nabla {f}(x^{(k)}; \xi^{(k)})}^2\right]\\
    &=\norm{x^{(k)}}^2-\frac{2}{{L}}\mathbb{E}\left[ \langle x^{(k)},  \nabla {f}(x^{(k)}; \xi^{(k)}) \rangle  \right]\\
    &+\frac{1}{{L}^2}\mathbb{E}\left[\norm{\nabla {f}(x^{(k)}; \xi^{(k)})}^2\right].
    \label{x+}
    \end{aligned}
\end{equation}
\newpage
Since 
\begin{equation}
    \begin{aligned}
        &\mathbb{E}\left[\norm{\nabla {f}(x^{(k)}; \xi^{(k)})}^2\right]\\
        =&\mathbb{E}\left[\norm{\nabla {f}(x^{(k)}; \xi^{(k)})-\nabla {f}(x^{(k)})+\nabla {f}(x^{(k)})}^2\right]\\
        =& \mathbb{E}\left[\norm{\nabla {f}(x^{(k)}; \xi^{(k)})-\nabla {f}(x^{(k)})}^2\right]+\norm{\nabla {f}(x^{(k)})}^2\\
        \leq& {\sigma}^2+\norm{\nabla {f}(x^{(k)})}^2.
        \label{avg}
    \end{aligned}
\end{equation}

Then plugging \eqref{prod} and \eqref{avg} to \eqref{x+}, we obtain
\begin{equation}
    \begin{aligned}
            &\mathbb{E}\left[\norm{x^{(k+1)}}^2\right] \\
    \leq&\norm{x^{(k)}}^2 +\frac{\norm{\nabla {f}(x^{(k)})}^2}{{L}^2}+\frac{{\sigma}^2}{{L}^2}\\
    &+\frac{2}{{L}}
    \left(\frac{{L}^{-1}||\nabla {f}(x^{(k)})||^2}{1+{\kappa}^{-1}} + \frac{\mu||x^{(k)}-x^\star||^2}{1+{\kappa}^{-1}}-r\norm{\nabla {f}(x^{(k)})}\right)\\
    =&\norm{x^{(k)}}^2+\frac{2r}{{L}}\norm{\nabla {f}(x^{(k)})}-\frac{1}{{L^2}}\left(\frac{1-{\kappa}^{-1}}{1+{\kappa}^{-1}}\right)\norm{\nabla {f}(x^{(k)})}^2 \\
    &- \frac{2 \mu}{L(1+{\kappa}^{-1})} \norm{x^{(k)}-x^\star}^2+\frac{{\sigma}^2}{{L}^2 } .
    \label{x+2}
    \end{aligned}
\end{equation}

Since ${\kappa}\geq 3$, we have that $\frac{1-{\kappa}^{-1}}{1+{\kappa}^{-1}}>0$. Thus the second term in \eqref{x+2} is strictly concave in $\norm{\nabla {f}(x^{(k)})}$, and hence can be bounded by its maximum which gives that
\begin{equation}
    \begin{aligned}
            \mathbb{E}\left[\norm{x^{(k+1)}}^2\right] 
    &\leq\norm{x^{(k)}}^2+r^2\frac{1+{\kappa}^{-1}}{1-{\kappa}^{-1}}-2\frac{{\kappa}^{-1}}{1+{\kappa}^{-1}}\norm{x^{(k)}-x^\star}^2+\frac{{\sigma}^2}{{L}^2 }\\
    &\leq\norm{x^{(k)}}^2+2r^2-{\kappa}^{-1}(\norm{x^{(k)}}-\norm{x^\star})^2+\frac{{\sigma}^2}{{L}^2 }.
    \label{x+3}
    \end{aligned}
\end{equation}

where the second inequality comes from the fact that $\frac{1+{\kappa}^{-1}}{1-{\kappa}^{-1}}\leq2$ and $-\frac{2{\kappa}^{-1}}{1+{\kappa}^{-1}}\leq-{\kappa}^{-1}$.

Then since $\norm{x^{(k)}}\geq r$, we have that $\norm{x^{(k)}}-r\geq 0$.
$$Q(\norm{x^{(k)}})=\norm{x^{(k)}}^2+2b^2-{\kappa}^{-1}(\norm{x^{(k)}}-r)^2+\frac{{\sigma}^2}{{L}^2 }
$$
Since $1-{\kappa}^{-1}>0$, $Q(\norm{x^{(k)}})$ is strongly convex in $\norm{x^{(k)}}$. Thus, $Q(\norm{x^{(k)}})$ achieves maximum when $\norm{x^{(k)}}= 0$ or $\norm{x^{(k)}}= R$. 
Observe that 

$$Q(R)=R^2, Q(0)<R^2.$$ Then $\mathbb{E}\left[\norm{x^{(k+1)}}^2\right]\leq R^2$, then we can arrive at the conclusion.

\end{proof}

\subsection{Proof of Lemma 4}
In this section, we present the detailed proof of Lemma 4.
\begin{proof}

From the L-smoothness of function $f$, we can get the following inequality:
\begin{equation}
    f(x^{(k+1)}) - f(x^{(k)}) \leq \frac{L}{2}\norm{x^{(k+1)} - x^{(k)}}^2 - \eta \underbrace{\langle \nabla f(x^{(k)}), (\epsilon I + \hat{V}^{(k+1)})^{-\frac{1}{2}} h^{(k+1)}\rangle}_{-A^{(k)}}
\end{equation}

For the ease of notation, we denote the term $-\langle \nabla f(x^{(k)}), (\epsilon I + \hat{V}^{(k+1)})^{-\frac{1}{2}} h^{(k+1)}\rangle$ as  $A^{(k)}$, next, we consider the upper bound for $A^{(k)}$, which can be decomposed into three parts denoted as $I_1, I_2$ and $I_3$ respectively.
\begin{equation}
    \begin{aligned}
            \mathbb{E}\left[A^{(k)}|x^{(k)}\right] =& -\mathbb{E}\left[\langle \nabla f(x^{(k)}), (\epsilon I + \hat{V}^{(k+1)})^{-\frac{1}{2}} h^{(k+1)}\rangle|x^{(k)}\right]\\
        =& -\mathbb{E}\left[\langle \nabla f(x^{(k)}), (\epsilon I + \hat{V}^{(k)})^{-\frac{1}{2}} h^{(k+1)}\rangle|x^{(k)}\right]\\
        &-\mathbb{E}\left[\langle \nabla f(x^{(k)}), [(\epsilon I + \hat{V}^{(k+1)})^{-\frac{1}{2}}-(\epsilon I + \hat{V}^{(k)})^{-\frac{1}{2}}] h^{(k+1)}\rangle|x^{(k)}\right]\\
        =& \underbrace{-\beta_1 \mathbb{E}\left[\langle \nabla f(x^{(k)}), (\epsilon I + \hat{V}^{(k)})^{-\frac{1}{2}} h^{(k)}\rangle|x^{(k)}\right]}_{I_1} \\
        &\underbrace{- (1-\beta_1)\mathbb{E}\left[\langle \nabla f(x^{(k)}), (\epsilon I + \hat{V}^{(k)})^{-\frac{1}{2}} \nabla f(x^{(k)}|\xi^{(k)})\rangle|x^{(k)}\right]}_{I_2}\\
        &\underbrace{-\mathbb{E}\left[\langle \nabla f(x^{(k)}), [(\epsilon I + \hat{V}^{(k+1)})^{-\frac{1}{2}}-(\epsilon I + \hat{V}^{(k)})^{-\frac{1}{2}}] h^{(k+1)}\rangle|x^{(k)}\right]}_{I_3}
    \end{aligned}
\end{equation}

Using properties from the functions, we can provide an upper bound for all the three terms above.

\begin{equation}
    \begin{aligned}
        I_1 =& -\beta_1 \mathbb{E}\left[\langle \nabla f(x^{(k)}), (\epsilon I + \hat{V}^{(k)})^{-\frac{1}{2}} h^{(k)}\rangle|x^{(k)}\right]\\
         =& -\beta_1 \mathbb{E}\left[\langle \nabla f(x^{(k-1)}), (\epsilon I + \hat{V}^{(k)})^{-\frac{1}{2}} h^{(k)}\rangle|x^{(k)}\right] \\
        &- \beta_1 \mathbb{E}\left[\langle \nabla f(x^{(k-1)}) - \nabla f(x^{(k)}), (\epsilon I + \hat{V}^{(k)})^{-\frac{1}{2}} h^{(k)}\rangle|x^{(k)}\right]\\
        \leq & \beta_1 A^{(k-1)} +\frac{\beta_1 L}{\eta}\norm{x^{(k)} - x^{(k-1)}}^2
    \end{aligned}
\end{equation}

\begin{equation}
    \begin{aligned}
        I_2 =& - (1-\beta_1)\mathbb{E}\left[\langle \nabla f(x^{(k)}), (\epsilon I + \hat{V}^{(k)})^{-\frac{1}{2}} \nabla f(x^{(k)}|\xi^{(k)})\rangle|x^{(k)}\right]\\
        =&- (1-\beta_1)\langle \nabla f(x^{(k)}), (\epsilon I + \hat{V}^{(k)})^{-\frac{1}{2}} \nabla f(x^{(k)})\rangle\\
        =&- (1-\beta_1) \norm{\nabla f(x^{(k)})}^2_{(\epsilon I + \hat{V}^{(k)})^{-\frac{1}{2}}}
    \end{aligned}
\end{equation}

\begin{equation}
    \begin{aligned}
        I_3 = & -\mathbb{E}\left[\langle \nabla f(x^{(k)}), [(\epsilon I + \hat{V}^{(k+1)})^{-\frac{1}{2}}-(\epsilon I + \hat{V}^{(k)})^{-\frac{1}{2}}] h^{(k+1)}\rangle|x^{(k)}\right]\\
        \leq & \mathbb{E}\left[ \norm{\nabla f(x^{(k)})}\norm{h^{(k+1)}}\norm{(\epsilon I + \hat{V}^{(k+1)})^{-\frac{1}{2}}-(\epsilon I + \hat{V}^{(k)})^{-\frac{1}{2}}}  |x^{(k)}\right]\\
        \leq & \sigma^2 \mathbb{E}\left[\sum_{i=1}^D [(\epsilon + [\hat{v}^{(k)}]_i)^{-\frac{1}{2}} - (\epsilon + [\hat{v}^{(k+1)}]_i)^{-\frac{1}{2}}] |x^{(k)}\right]
    \end{aligned}
\end{equation}

Then we can write the relationship of Lyapunov function between two consecutive iterations.

\begin{equation}
\resizebox{\hsize}{!}{$
    \begin{aligned}
        \mathbb{E}\left[\mathcal{L}^{(k+1)}\right] - \mathcal{L}^{(k)}  = & \mathbb{E}\left[f(x^{(k+1)}) - c \langle \nabla f(x^{(k)}), (\epsilon I + \hat{V}^{(k+1)})^{-\frac{1}{2}} h^{(k+1)}\rangle  + b \sum_{i=1}^d (\epsilon + [\hat{v}^{(k+1)}]_i)^{-\frac{1}{2}}- f(x^{(k)})\right]\\
    & + c \langle \nabla f(x^{(k-1)}), (\epsilon I + \hat{V}^{(k)})^{-\frac{1}{2}} h^{(k)}\rangle  - b \sum_{i=1}^d (\epsilon + [\hat{v}^{(k)}]_i)^{-\frac{1}{2}})\\
    =&f(x^{(k+1)}) - f(x^{(k)}) + c \mathbb{E}\left[A^{(k)}\right] - c A^{(k-1)}  + b \mathbb{E}\left[\sum_{i=1}^d (\epsilon + [\hat{v}^{(k+1)}]_i)^{-\frac{1}{2}}- \sum_{i=1}^d (\epsilon + [\hat{v}^{(k)}]_i)^{-\frac{1}{2}})\right]\\
    \leq & \frac{L}{2}\norm{x^{(k+1)} - x^{(k)}}^2 + (\eta+c) \mathbb{E}\left[A^{(k)}\right]  - c A^{(k-1)} + b \mathbb{E}\left[\sum_{i=1}^d (\epsilon + [\hat{v}^{(k+1)}]_i)^{-\frac{1}{2}}- \sum_{i=1}^d (\epsilon + [\hat{v}^{(k)}]_i)^{-\frac{1}{2}})\right]\\
    \leq & \frac{L}{2}\norm{x^{(k+1)} - x^{(k)}}^2  - c A^{(k-1)}  + b \mathbb{E}\left[\sum_{i=1}^d (\epsilon + [\hat{v}^{(k+1)}]_i)^{-\frac{1}{2}}- \sum_{i=1}^d (\epsilon + [\hat{v}^{(k)}]_i)^{-\frac{1}{2}})\right]\\
    &+ (\eta + c)(\beta_1 A^{(k-1)} +\frac{\beta_1 L}{\eta}\norm{x^{(k)} - x^{(k-1)}}^2  - (1-\beta_1) \norm{\nabla f(x^{(k)})}^2_{(\epsilon I + \hat{V}^{(k)})^{-\frac{1}{2}}} \\
    &+ \sigma^2 \mathbb{E}\left[\sum_{i=1}^d [(\epsilon + [\hat{v}^{(k)}]_i)^{-\frac{1}{2}} - (\epsilon + [\hat{v}^{(k+1)}]_i)^{-\frac{1}{2}}] |x^{(k)}\right]  )
    \end{aligned}
    $}
\end{equation}

Now, if we set $$b = \sigma^2((\eta + c)), (\eta + c)\beta_1 = c,$$ i.e. $$c = \frac{\eta \beta_1}{1 - \beta_1}, b = \frac{\eta \sigma^2}{1-\beta_1}$$ then the equation above can be greatly simplified.
\begin{equation}
    \begin{aligned}
    \mathbb{E}\left[\mathcal{L}^{(k+1)}\right]   \leq & \mathcal{L}^{(k)} + (\frac{L}{2}+ (\eta + c)\frac{\beta_1 L}{\eta})\norm{x^{(k+1)} - x^{(k)}}^2  - (\eta+ c) (1-\beta_1) \norm{\nabla f(x^{(k)})}^2_{(\epsilon I + \hat{V}^{(k)})^{-\frac{1}{2}}} \\
    \leq &\mathcal{L}^{(k)} -(\eta + c) (1-\beta_1) \norm{\nabla f(x^{(k)})}^2_{(\epsilon I + \hat{V}^{(k)})^{-\frac{1}{2}}}  + (\frac{L}{2}+ \frac{ L}{1-\beta_1})\norm{x^{(k+1)} - x^{(k)}}^2\\
    \leq & \mathcal{L}^{(k)}-\eta  (\epsilon + \frac{\sigma^2}{1-\beta_2})^{-\frac{1}{2}}\norm{\nabla f(x^{(k)})}^2  + (\frac{L}{2}+ \frac{ L}{1-\beta_1})\norm{x^{(k+1)} - x^{(k)}}^2\\
    \leq &\mathcal{L}^{(k)} -2 \mu\eta (\epsilon + \frac{\sigma^2}{1-\beta_2})^{-\frac{1}{2}}(f(x^{(k)}) - f(x^\star))  + (\frac{L}{2}+ \frac{ L}{1-\beta_1})\norm{x^{(k+1)} - x^{(k)}}^2\\
    \leq & \mathcal{L}^{(k)}-2 \mu\eta  (\epsilon + \frac{\sigma^2}{1-\beta_2})^{-\frac{1}{2}}(\mathcal{L}^{(k)} - b \sum_{i=1}^D (\epsilon + [\hat{v}^{(k)}]_i)^{-\frac{1}{2}} + c \langle \nabla f(x^{(k-1)}), (\epsilon I + \hat{V}^{(k)})^{-\frac{1}{2}} h^{(k)}\rangle ) \\
    & + (\frac{L}{2}+ \frac{ L}{1-\beta_1})\norm{x^{(k+1)} - x^{(k)}}^2\\
    =& (1 - 2 \mu\eta (\epsilon + \frac{\sigma^2}{1-\beta_2})^{-\frac{1}{2}}) \mathcal{L}^{(k)}+ (\frac{L}{2}+ \frac{ L}{1-\beta_1})\norm{x^{(k+1)} - x^{(k)}}^2\\
    &  -2 \mu\eta  (\epsilon + \frac{\sigma^2}{1-\beta_2})^{-\frac{1}{2}} (- b \sum_{i=1}^D (\epsilon + [\hat{v}^{(k)}]_i)^{-\frac{1}{2}} +c \langle \nabla f(x^{(k-1)}), (\epsilon I + \hat{V}^{(k)})^{-\frac{1}{2}} h^{(k)}\rangle  )\\
    \leq & (1 - 2 \mu\eta  (\epsilon + \frac{\sigma^2}{1-\beta_2})^{-\frac{1}{2}}) \mathcal{L}^{(k)}+ (\frac{L}{2}+ \frac{ L}{1-\beta_1})\frac{\eta^2 D}{(1-\beta_2)(1 - \beta_1^2/\beta_2)}\\
    &  +2 \mu\eta^2  (\epsilon + \frac{\sigma^2}{1-\beta_2})^{-\frac{1}{2}} \frac{ \beta_1\sigma^2 \epsilon^{-\frac{1}{2}}}{1 - \beta_1}+ 2 \mu\eta^2  (\epsilon + \frac{\sigma^2}{1-\beta_2})^{-\frac{1}{2}} \frac{ \sigma^2 D \epsilon ^{-\frac{1}{2}}}{1-\beta_1} \\
    =& C_1 V_k + C_2
    \end{aligned}
\end{equation}
The last inequality is based on Lemma 8 in \cite{chen2021cada}.
\end{proof}

        

\section{ADDITIONAL EXPERIMENTS}

In addition to the convex logistic loss, we conducted experiments with convolutional neural networks (CNNs) as client models, in the same open FL system setting described in Section 4. Due to the choice of client models, the landscape of the local objective functions  in this case is not guaranteed to be convex. The CNN model at each client consist of two convolutional layers, each followed by a ReLU activation layer and a max pooling layer. 

We distribute the data among the clients independently and identically such that each client have $m=100$ data. The system consist of a pool of $N=600$ clients, and in each simulation the current number of clients initialize at $N_0=10$. The local loss function at each client is the $\ell_2$ regularized cross-entropy.Similar to the experiments described in Section 4, we investigate the effect of the regularization parameter $\lambda$ of the local loss function and the probability $p$ of clients joining and/or leaving the system. We provide results for local Adam, since the behavior of local SGD and local Adam are similar. The local optimization parameters of clients are set as $\beta_1=0.9$, $\beta_2=0.999$, $\epsilon=10^{-8}$ and $\eta=0.005$.


\begin{figure}[h!]
\centering
\begin{subfigure}{0.45\textwidth}
  \centering
  \includegraphics[width=1\linewidth]{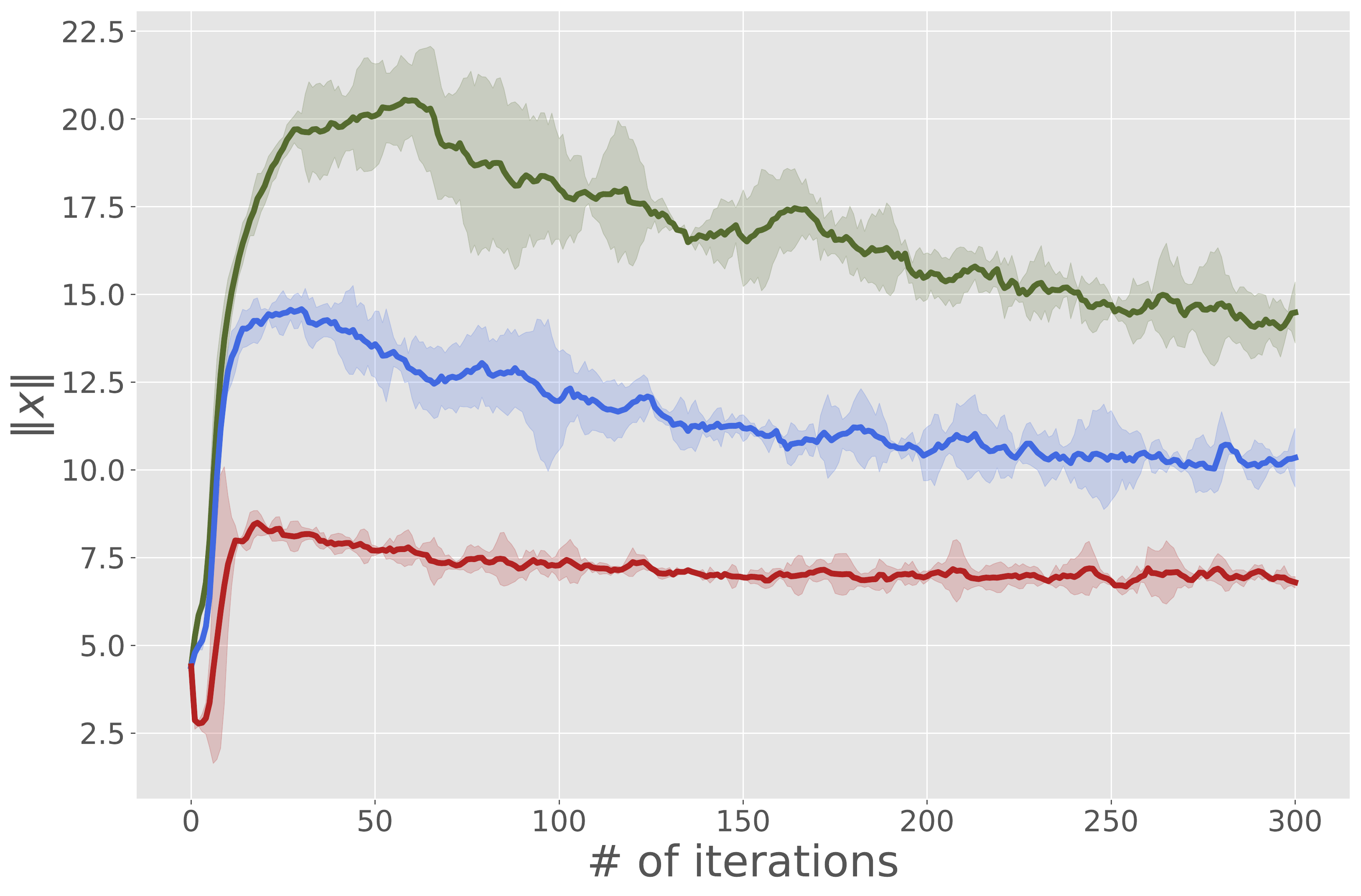}
  \caption{Iterate norm}
  \label{fig:sub-first}
\end{subfigure}
\begin{subfigure}{0.45\textwidth}
  \centering
  \includegraphics[width=1\linewidth]{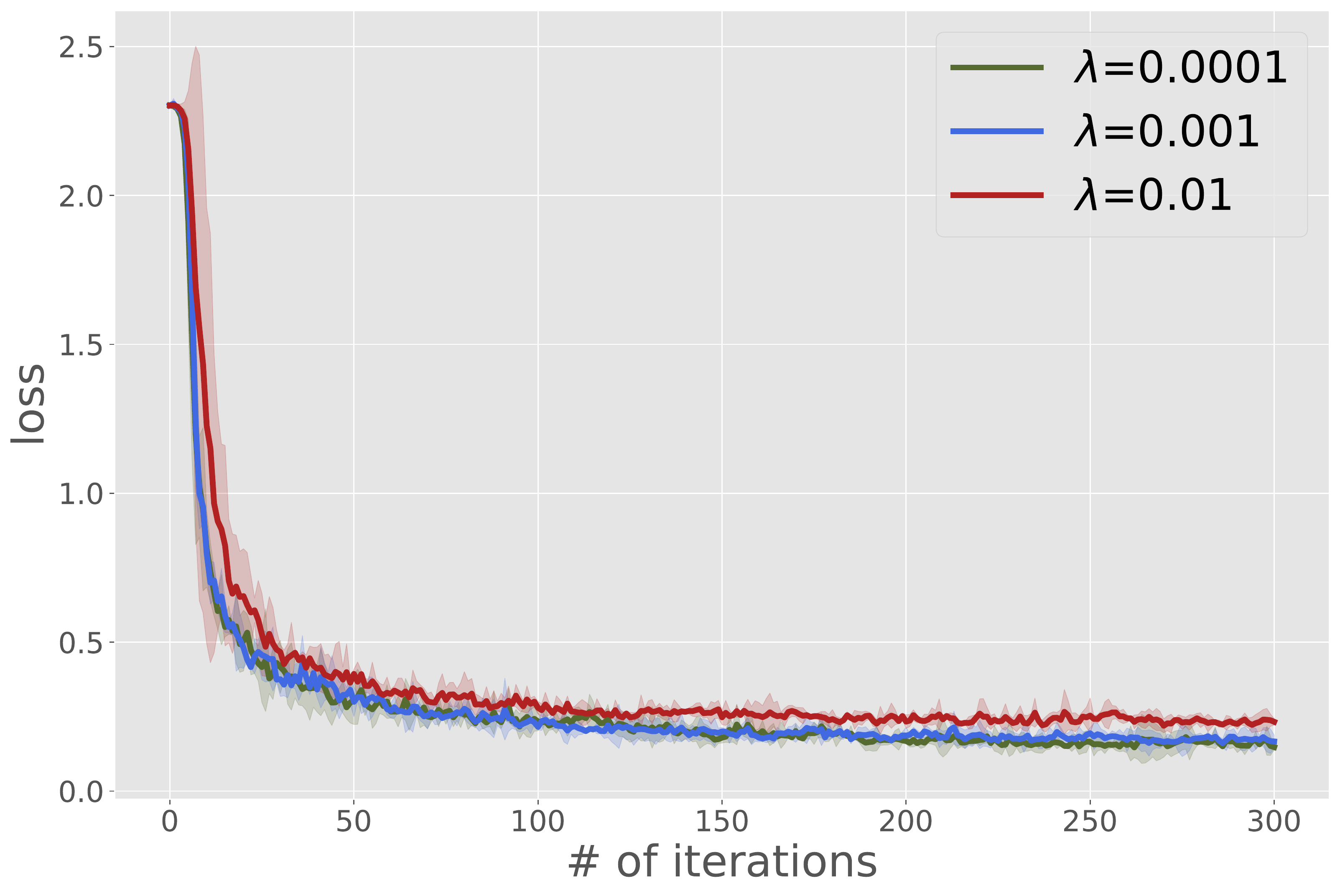}
  \caption{Loss}
  \label{fig:sub-second}
\end{subfigure}
\caption{Comparison of iterate norms and loss for different $\lambda$ for local Adam.} 
\label{fig:mnist-cnn-adam-lambda}
\vspace{-0.2cm}
\end{figure}


\begin{figure}[h!]
\centering
\begin{subfigure}{0.45\textwidth}
  \centering
  \includegraphics[width=1\linewidth]{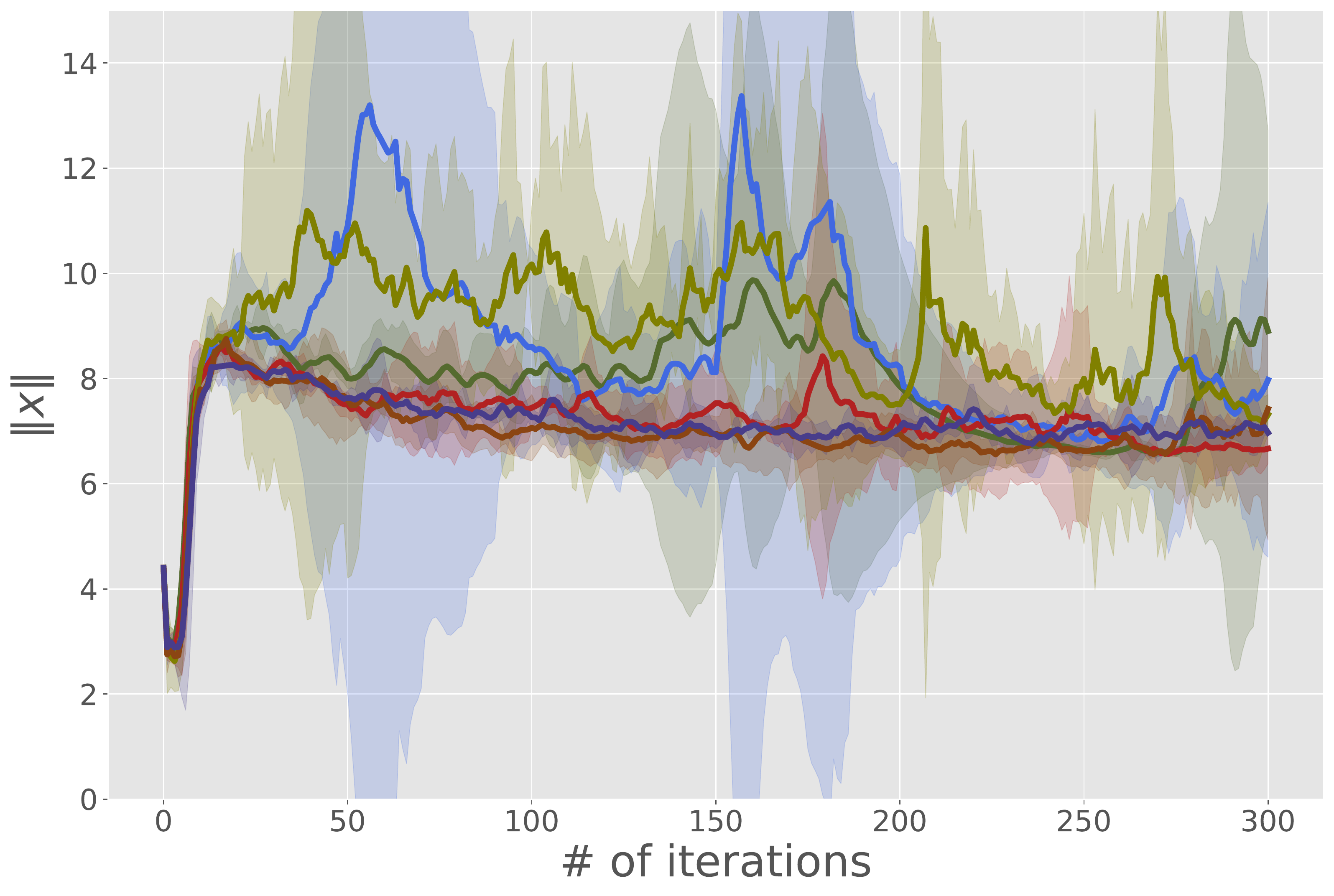}
  \caption{Iterate norm}
  \label{fig:sub-first}
\end{subfigure}
\begin{subfigure}{0.45\textwidth}
  \centering
  \includegraphics[width=1\linewidth]{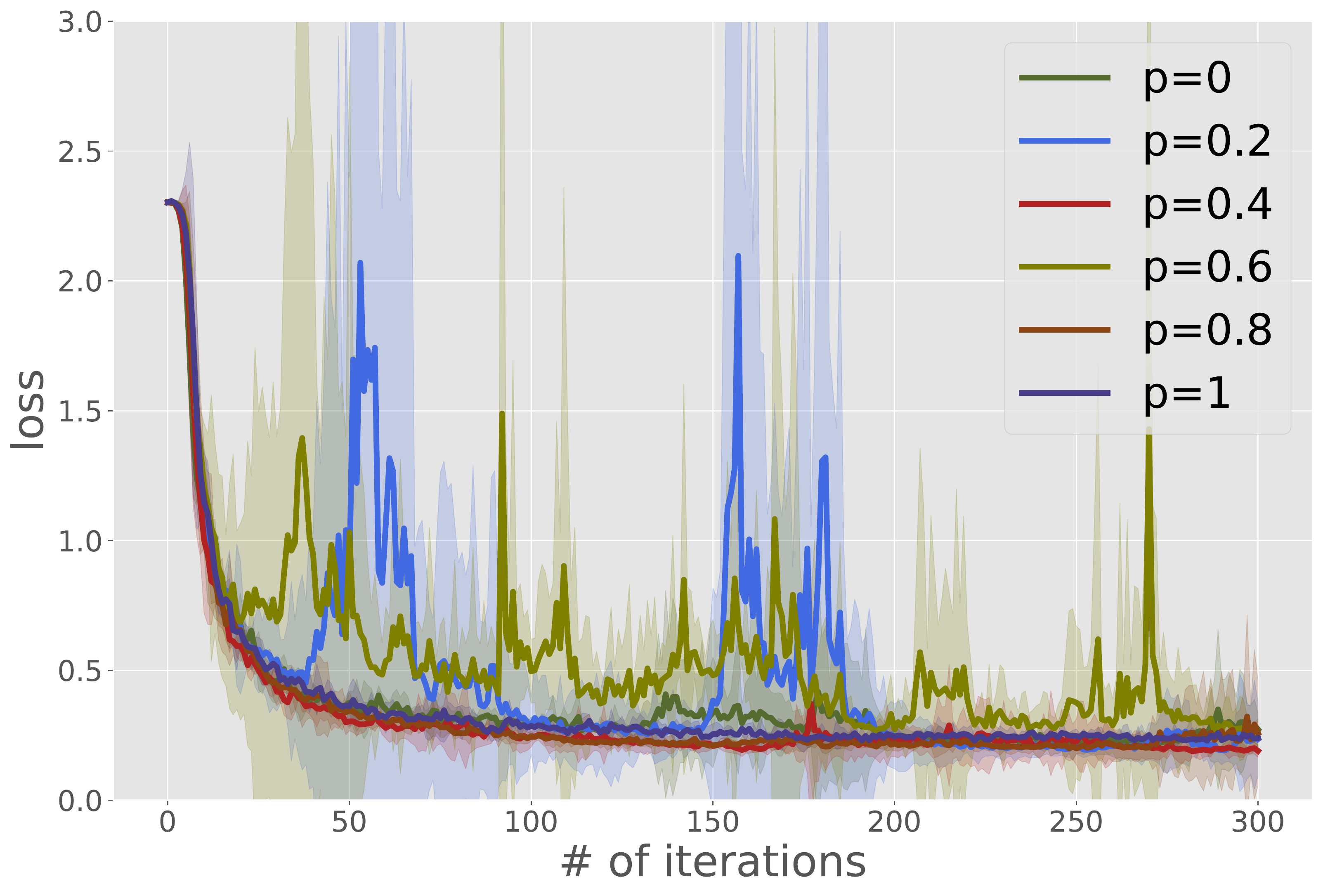}
  \caption{Loss}
  \label{fig:sub-second}
\end{subfigure}
\caption{Comparison of iterate norms and loss for different $p$ for local Adam.} 
\label{fig:mnist-cnn-adam-p}
\vspace{-0.2cm}
\end{figure}

Figure \ref{fig:mnist-cnn-adam-lambda} shows how the global iterate norm  and the global loss change with the regularization parameter $\lambda$ for local Adam, for $p=1$. It can be seen that the global iterate norm converge to a stable value after some iterations for the tested $\lambda$ values, even though the local models are not guaranteed to be convex. As expected the stable value of global iterate norm is higher when $\lambda$ is smaller. 

Figure \ref{fig:mnist-cnn-adam-p} show how the global iterate norm and the global loss behave with changing $p$ for local Adam. The global loss for different $p$ differs only slightly, but shows high variance in some cases due to the dynamic nature of the system. Accordingly, the global iterate seem to achieve stability after some number of iterations for all tested $p$. These empirical results suggest that non-convex models like CNNs show stability in open FL systems.



\bibliographystyle{apalike}

\bibliography{ref}

%% file: header.tex
\newcommand\numberthis{\addtocounter{equation}{1}\tag{\theequation}}




\newcommand{\0}{\mathbb{0}}
\newcommand{\1}{\mathbb{1}}

\newcommand{\E}{\mathbb{E}}
\newcommand{\R}{\mathbb{R}}
\renewcommand{\P}{\mathbb{P}}
\newcommand{\U}{\mathbb{U}}

\newcommand{\diag}{\text{diag}}
\newcommand{\col}{\text{col}}


\newcommand{\ab}{\mathbf{a}}
\newcommand{\bb}{\mathbf{b}}
\newcommand{\cb}{\mathbf{c}}
\newcommand{\db}{\mathbf{d}}
\newcommand{\eb}{\mathbf{e}}
\newcommand{\fb}{\mathbf{f}}
\newcommand{\gb}{\mathbf{g}}
\newcommand{\ib}{\mathbf{i}}
\newcommand{\kb}{\mathbf{k}}
\newcommand{\pb}{\mathbf{p}}
\newcommand{\qb}{\mathbf{q}}
\newcommand{\vb}{\mathbf{v}}
\newcommand{\ub}{\mathbf{u}}
\newcommand{\wb}{\mathbf{w}}
\newcommand{\xb}{\mathbf{x}}
\newcommand{\yb}{\mathbf{y}}
\newcommand{\zb}{\mathbf{z}}

\newcommand{\Ab}{\mathbf{A}}
\newcommand{\Bb}{\mathbf{B}}
\newcommand{\Cb}{\mathbf{C}}
\newcommand{\Db}{\mathbf{D}}
\newcommand{\Eb}{\mathbf{E}}
\newcommand{\Fb}{\mathbf{F}}
\newcommand{\Gb}{\mathbf{G}}
\newcommand{\Hb}{\mathbf{H}}
\newcommand{\Ib}{\mathbf{I}}
\newcommand{\Jb}{\mathbf{J}}
\newcommand{\Kb}{\mathbf{K}}
\newcommand{\Lb}{\mathbf{L}}
\newcommand{\Mb}{\mathbf{M}}
\newcommand{\Pb}{\mathbf{P}}
\newcommand{\Qb}{\mathbf{Q}}
\newcommand{\Rb}{\mathbf{R}}
\newcommand{\Sb}{\mathbf{S}}
\newcommand{\Vb}{\mathbf{V}}
\newcommand{\Ub}{\mathbf{U}}
\newcommand{\Wb}{\mathbf{W}}
\newcommand{\Xb}{\mathbf{X}}
\newcommand{\Yb}{\mathbf{Y}}
\newcommand{\Zb}{\mathbf{Z}}

\newcommand{\alphab}{\boldsymbol{\alpha}}
\newcommand{\betab}{\boldsymbol{\beta}}
\newcommand{\gammab}{\boldsymbol{\gamma}}
\newcommand{\phib}{\boldsymbol{\phi}}
\newcommand{\Phib}{\boldsymbol{\Phi}}
\newcommand{\Qhib}{\boldsymbol{\Qhi}}
\newcommand{\omegab}{\boldsymbol{\omega}}
\newcommand{\psib}{\boldsymbol{\psi}}
\newcommand{\sigmab}{\boldsymbol{\sigma}}
\newcommand{\nub}{\boldsymbol{\nu}}
\newcommand{\thetab}{\boldsymbol{\theta}}
\newcommand{\delb}{\boldsymbol{\delta}}
\newcommand{\rhob}{\boldsymbol{\rho}}
\newcommand{\Pib}{\boldsymbol{\Pi}}
\newcommand{\pib}{\boldsymbol{\pi}}
\newcommand{\Sigmab}{\boldsymbol{\Sigma}}

\newcommand{\Ac}{\mathcal{A}}
\newcommand{\Bc}{\mathcal{B}}
\newcommand{\Cc}{\mathcal{C}}
\newcommand{\Ec}{\mathcal{E}}
\newcommand{\Fc}{\mathcal{F}}
\newcommand{\Hc}{\mathcal{H}}
\newcommand{\Ic}{\mathcal{I}}
\newcommand{\Lc}{\mathcal{L}}
\newcommand{\Nc}{\mathcal{N}}
\newcommand{\Oc}{\mathcal{O}}
\newcommand{\Pc}{\mathcal{P}}
\newcommand{\Rc}{\mathcal{R}}
\newcommand{\Sc}{\mathcal{S}}
\newcommand{\Uc}{\mathcal{U}}
\newcommand{\Vc}{\mathcal{V}}
\newcommand{\Xc}{\mathcal{X}}
\newcommand{\Yc}{\mathcal{Y}}


\newcommand{\tauh}{\widehat{\tau}}
\newcommand{\Sigmah}{\widehat{\Sigma}}

\newcommand{\fh}{\widehat{f}}
\newcommand{\gh}{\widehat{g}}
\newcommand{\kh}{\widehat{k}}
\newcommand{\qh}{\widehat{q}}
\newcommand{\Rh}{\widehat{R}}


\newcommand{\alphabh}{\widehat{\boldsymbol{\alpha}}}
\newcommand{\thetabh}{\widehat{\boldsymbol{\theta}}}

\newcommand{\qbh}{\widehat{\mathbf{q}}}

\newcommand{\Kbh}{\widehat{\mathbf{K}}}


\newcommand{\Fch}{\widehat{\mathcal{F}}}


\newcommand{\argmin}{\text{argmin}}
\newcommand{\arginf}{\text{arginf}}
\newcommand{\argmax}{\text{argmax}}
\newcommand{\minimize}{\text{minimize}}
\newcommand{\maximize}{\text{maximize}}
\newcommand{\supp}{\text{supp}}


\newcommand{\TV}{\text{TV}}
\newcommand{\norm}[1]{\lVert#1\rVert}
\newcommand{\tr}[1]{\text{Tr}\left[#1\right]}
\newcommand{\inn}[1]{\left<#1\right>}
\newcommand{\seal}[1]{\left \lceil #1\right \rceil}
\newcommand{\floor}[1]{\left \lfloor #1\right \rfloor}
\newcommand{\abs}[1]{\left|#1\right|}
\newcommand{\ind}[1]{\mathbf{1}\left(#1\right)}
\newcommand{\ex}[1]{\E\left[#1\right]}


\newtheorem{theorem}{Theorem}
\newtheorem{acknowledgement}[theorem]{Acknowledgement}
\newtheorem{assumption}{Assumption}
\newtheorem{conjecture}[theorem]{Conjecture}
\newtheorem{corollary}[theorem]{Corollary}
\newtheorem{definition}{Definition}
\newtheorem{example}{Example}
\newtheorem{lemma}[theorem]{Lemma}
\newtheorem{fact}{Fact}
\newtheorem{problem}{Problem}
\newtheorem{proposition}[theorem]{Proposition}
\newtheorem{remark}{Remark}
\newtheorem{solution}[theorem]{Solution}
\newtheorem{summary}[theorem]{Summary}

\newcommand{\bl}{\color{blue}}
\newcommand{\rd}{\color{red}}

%% file: supp_content.tex
\onecolumn
\section{Omitted Proofs}
We now provide the proof for theorems and related lemmas presented in the paper.
\subsection{Proof of Theorem 2}
We first introduce the following lemma to replace the arbitrary learning rate $\eta$ with a fixed learning rate $\frac{1}{L}$.

\begin{lemma}\label{lemma_eta}
    If for variables $x, \Delta x$, there exists some finite scalar $R$ that satisfies $\norm{x} \leq R$ and $\norm{x - \frac{1}{L} \Delta x} \leq R$, then $\norm{x - \eta \Delta x} \leq R$ is true for any $\eta \in (0, \frac{1}{L}]$.
\end{lemma}
\begin{proof}
Since ${x} \in \mathbf{B}(0, R)$ and ${x - \frac{1}{L} \Delta x} \in \mathbf{B}(0, R)$, and that $\mathbf{B}(0, R)$ is a convex set, ${x - \eta \Delta x} $ which is a convex combination of the former two will also be in the same convex set.
\end{proof}
We can replace the learning rate in SGD with the maximum possible learning rate $\frac{1}{L}$ using Lemma \ref{lemma_eta}. Next, we present our proof for Theorem 2.
\begin{proof}
We first assume that the condition number $\kappa  = \frac{L}{\mu} \geq 3$; note that when the function has a better condition number (i.e. $\kappa < 3$), we consider its condition number as equal to $3$, since we can consider a smaller number $\mu' < \mu$, and a $\mu$-strongly convex function is always $\mu'$-strongly convex.

We consider the $k$-th iteration. From the strong convexity and smoothness assumptions of the objective function, we can write the following inequality,
\begin{equation}
    \begin{aligned}
        &\mathbb{E}\left[ \langle x^{(k)} - x^\star,  \nabla {f}(x^{(k)}; \xi^{(k)}) \rangle\big|x^{(k)}\right] = \langle x^{(k)} - x^\star, \nabla {f}(x^{(k)})\rangle\geq \frac{{L}^{-1}\norm{\nabla {f}(x^{(k)})}^2}{1+{\kappa}^{-1}} + \frac{\mu \norm{x^{(k)}-x^\star}^2}{1+{\kappa}^{-1}}
    \end{aligned}
\end{equation}
Adding $\mathbb{E}\left[ \langle \nabla {f}(x^{(k)}; \xi^{(k)}) , x^\star \rangle \big|x^{(k)} \right]$ to both sides, using Cauchy-Schwartz inequality and the assumption that $\norm{x^\star}\leq r$, we obtain
\begin{equation}
    \begin{aligned}
        &\mathbb{E}\left[  \langle x^{(k)},  \nabla {f}(x^{(k)}; \xi^{(k)}) \rangle \big|x^{(k)}\right]\geq \frac{{L}^{-1}\norm{\nabla {f}(x^{(k)})}^2}{1+{\kappa}^{-1}} + \frac{\mu \norm{x^{(k)}-x^\star}^2}{1+{\kappa}^{-1}}-r \norm{\nabla {f}(x^{(k)})}.
    \label{prod}
    \end{aligned}
\end{equation}
We focus on the case $\eta=\frac{1}{{L}}$. By the definition of SGD,
\begin{equation}
    \begin{aligned}
    &\mathbb{E}\left[\norm{x^{(k+1)}}^2\big|x^{(k)}\right] = \mathbb{E}\left[\norm{x^{(k)} - \frac{1}{{L}}\nabla {f}(x^{(k)}; \xi^{(k)})}^2\Big|x^{(k)}\right]\\
    =&\norm{x^{(k)}}^2-\frac{2}{{L}}\mathbb{E}\left[ \langle x^{(k)},  \nabla {f}(x^{(k)}; \xi^{(k)}) \rangle \Big|x^{(k)} \right]+\frac{1}{{L}^2}\mathbb{E}\left[\norm{\nabla {f}(x^{(k)}; \xi^{(k)})}^2\Big|x^{(k)}\right].
    \label{x+}
    \end{aligned}
\end{equation}
Since 
\begin{equation}
    \begin{aligned}
        \mathbb{E}\left[\norm{\nabla {f}(x^{(k)}; \xi^{(k)})}^2\big|x^{(k)}\right]=&\mathbb{E}\left[\norm{\nabla {f}(x^{(k)}; \xi^{(k)})-\nabla {f}(x^{(k)})+\nabla {f}(x^{(k)})}^2\big|x^{(k)}\right]\\
        =& \mathbb{E}\left[\norm{\nabla {f}(x^{(k)}; \xi^{(k)})-\nabla {f}(x^{(k)})}^2\big|x^{(k)}\right]+\norm{\nabla {f}(x^{(k)})}^2\leq {\sigma}^2+\norm{\nabla {f}(x^{(k)})}^2.
        \label{avg}
    \end{aligned}
\end{equation}
Then plugging \eqref{prod} and \eqref{avg} to \eqref{x+}, we obtain
\begin{equation}
    \begin{aligned}
            \mathbb{E}\left[\norm{x^{(k+1)}}^2\big|x^{(k)}\right] 
    \leq&\norm{x^{(k)}}^2 +\frac{\norm{\nabla {f}(x^{(k)})}^2}{{L}^2}+\frac{{\sigma}^2}{{L}^2}\\
    &-\frac{2}{{L}}
    \left(\frac{{L}^{-1}\norm{\nabla {f}(x^{(k)})}^2}{1+{\kappa}^{-1}} + \frac{\mu\norm{x^{(k)}-x^\star}^2}{1+{\kappa}^{-1}}-r\norm{\nabla {f}(x^{(k)})}\right)\\
    =&\norm{x^{(k)}}^2+\frac{2r}{{L}}\norm{\nabla {f}(x^{(k)})}-\frac{1}{{L^2}}\left(\frac{1-{\kappa}^{-1}}{1+{\kappa}^{-1}}\right)\norm{\nabla {f}(x^{(k)})}^2 \\
    & -\frac{2 \mu}{L(1+{\kappa}^{-1})} \norm{x^{(k)}-x^\star}^2+\frac{{\sigma}^2}{{L}^2 } .
    \label{x+2}
    \end{aligned}
\end{equation}

Since ${\kappa}\geq 3$, we have that $\frac{1-{\kappa}^{-1}}{1+{\kappa}^{-1}}>0$. Thus, \eqref{x+2} is strictly concave in $\norm{\nabla {f}(x^{(k)})}$, and hence it can be bounded by its maximum, which implies
\begin{equation}
    \begin{aligned}
            \mathbb{E}\left[\norm{x^{(k+1)}}^2\big|x^{(k)}\right] 
    &\leq\norm{x^{(k)}}^2+r^2\frac{1+{\kappa}^{-1}}{1-{\kappa}^{-1}}-2\frac{{\kappa}^{-1}}{1+{\kappa}^{-1}}\norm{x^{(k)}-x^\star}^2+\frac{{\sigma}^2}{{L}^2 }\\
    &\leq\norm{x^{(k)}}^2+2r^2-{\kappa}^{-1}(\norm{x^{(k)}}-\norm{x^\star})^2+\frac{{\sigma}^2}{{L}^2 },
    \label{x+3}
    \end{aligned}
\end{equation}
where the second inequality comes from the fact that $\frac{1+{\kappa}^{-1}}{1-{\kappa}^{-1}}\leq2$ and $-\frac{2{\kappa}^{-1}}{1+{\kappa}^{-1}}\leq-{\kappa}^{-1}$.

When $\norm{x^{(k)}}< r$ the problem becomes trivial. For the case of $\norm{x^{(k)}}\geq r$, we have that $\norm{x^{(k)}}-r\geq 0$, and 
$$Q(\norm{x^{(k)}})=\norm{x^{(k)}}^2+2r^2-{\kappa}^{-1}(\norm{x^{(k)}}-r)^2+\frac{{\sigma}^2}{{L}^2}.
$$
Since $1-{\kappa}^{-1}>0$, $Q(\norm{x^{(k)}})$ is strongly convex in $\norm{x^{(k)}}$. Thus, $Q(\norm{x^{(k)}})$ achieves maximum when $\norm{x^{(k)}}= 0$ or $\norm{x^{(k)}}= R$. 
Observe that 

$$Q(R)=R^2, Q(0)<R^2.$$ Then $\mathbb{E}\left[\norm{x^{(k+1)}}^2\big|x^{(k)}\right]\leq R^2$, and we  arrive at the conclusion.

\end{proof}

\subsection{Proof of Theorem 3}\label{sec:thm}

First, we provide the explicit forms of constants $C_1-C_5$ defined as follows,
\begin{equation}
    \begin{aligned}
        C_1 = & 1 - 2 \mu\eta  (\epsilon + \sigma^2)^{-\frac{1}{2}}\\
        C_2 = & 2 \mu\eta^2  (\epsilon + \sigma^2)^{-\frac{1}{2}}  \sigma^2 d \epsilon^{-\frac{1}{2}}\left(\frac{1+\beta_1}{1-\beta_1}\right)
         + (\frac{L}{2}+ \frac{ L\beta_1}{1-\beta_1})\frac{\eta^2 d}{(1-\beta_2)(1 - \beta_1^2/\beta_2)}\\
        C_3 = &  \frac{\eta \beta_1}{1 - \beta_1} \sigma^2 \epsilon^{-\frac{1}{2}} - \frac{\eta \sigma^2}{1-\beta_1} d (\epsilon + \sigma^2)^{-\frac{1}{2}}\\
        C_4 = &  \frac{\eta \beta_1}{1 - \beta_1} \sigma^2 \epsilon^{-\frac{1}{2}} + \frac{\eta \sigma^2}{1-\beta_1} d \epsilon^{-\frac{1}{2}}\\
        C_5 = & \frac{1 + \kappa C_1}{1 - \kappa C_1}.
    \end{aligned}
\end{equation}
Note that $\epsilon$ and $\sigma$ should be such that $1-\kappa C_1>0$, which imposes a constraint on $\sigma$. For the Adam algorithm, since the update step is scaled in an adaptive fashion, we choose a slightly different approach for the analysis, compared to SGD. Inspired by \cite{chen2021cada}, we construct a Lyapunov function at step $k$ as follows
\begin{align}\label{eq:lyp}
    \mathcal{L}^{(k)} &= f(x^{(k)}) - f(x^\star)  + b \sum_{m=1}^d (\epsilon + [\hat{v}^{(k)}]_m)^{-\frac{1}{2}}  -c\langle \nabla f(x^{(k-1)}), (\epsilon I + \hat{V}^{(k)})^{-\frac{1}{2}} h^{(k)}\rangle,
\end{align}
where $b$ and $c$ are positive scalars, chosen appropriately in the analysis. The expression above is a valid choice for Lyapunov function since the function is uniformly lower bounded, and it can be shifted up by adding a positive constant.

We propose the following lemma on the contraction of Lyapunov function, which is crucial to derive the stability result in Theorem \ref{thm_adam}.

\begin{lemma} \label{lemma_lya}
    The (conditional) expectation of the Lyapunov function in \eqref{eq:lyp} is a contraction, i.e., there exists constants $C_1 \in (0,1), C_2\in \mathbb{R}^+$, such that the Lyapunov function satisfies the following inequality:
\begin{align}\label{eq:10}
    \mathbb{E}\left[\mathcal{L}^{(k+1)}\big|x^{(k)}\right] \leq C_1 \mathcal{L}^{(k)} + C_2.
\end{align}
\end{lemma}
The proof of lemma appears later in Section \ref{sec:lemma}. Since stability is defined with respect to optimization iterates, we need to provide upper and lower bounds for $\mathcal{L}^{(k)}$ in terms of $\norm{x^{(k)} - x^\star}$. From the definition of $\mathcal{L}^{(k)}$ in \eqref{eq:lyp} and Assumption \ref{adam_grad_assumption}, we have that
\begin{equation}
    \begin{aligned}
        f(x^{(k)}) - f(x^\star) =& \mathcal{L}^{(k)} + c \langle \nabla f(x^{(k-1)}), (\epsilon I + \hat{V}^{(k)})^{-\frac{1}{2}} h^{(k)}\rangle \nonumber\\
        & - b \sum_{m=1}^d (\epsilon + [\hat{v}^{(k)}]_m)^{-\frac{1}{2}}\\
        \leq & \mathcal{L}^{(k)} + c \sigma^2 \epsilon^{-\frac{1}{2}} - b d (\epsilon + \sigma^2)^{-\frac{1}{2}}\\
        = & \mathcal{L}^{(k)} + C_3.
    \end{aligned}
\end{equation}
By the strong convexity of function $f$, we can write 
\begin{equation}
    \begin{aligned}\label{eq:15}
        \norm{x^{(k)} - x^\star}^2 \leq &\frac{2}{\mu}\left( f(x^{(k)}) - f(x^\star)\right)
        \leq  \frac{2}{\mu} (\mathcal{L}^{(k)} + C_3).
    \end{aligned}
\end{equation}
Similarly, we can write
\begin{equation}
    \begin{aligned}
        f(x^{(k)}) - f(x^\star) =& \mathcal{L}^{(k)} + c \langle \nabla f(x^{(k-1)}), (\epsilon I + \hat{V}^{(k)})^{-\frac{1}{2}} h^{(k)}\rangle \nonumber\\
        & - b \sum_{m=1}^d (\epsilon + [\hat{v}^{(k)}]_m)^{-\frac{1}{2}}\\
        \geq & \mathcal{L}^{(k)} - c \sigma^2 \epsilon^{-\frac{1}{2}} - bd \epsilon^{-\frac{1}{2}}\\
        = & \mathcal{L}^{(k)} - C_4.
    \end{aligned}
\end{equation}
By the Lipschitz smoothness of function $f$, we can write
\begin{equation}\label{eq:20}
    \begin{aligned}
        \norm{x^{(k)} - x^\star}^2 \geq \frac{2}{L} \left( f(x^{(k)}) - f(x^\star)\right)
        \geq  \frac{2}{L} (\mathcal{L}^{(k)} -C_4).
    \end{aligned}
\end{equation}
Combining \eqref{eq:10}-\eqref{eq:15}-\eqref{eq:20}, we get
\begin{equation}
    \begin{aligned}
        \mathbb{E}\left[\norm{x^{(k+1)} - x^\star}^2\Big|x^{(k)} \right] \leq \frac{L}{\mu}C_1 \norm{x^{(k)} - x^\star}^2 + \frac{2}{\mu}(C_2 + C_3+ C_1 C_4).
    \end{aligned}
\end{equation}
Given that the objective function has an optimal solution $x^\star \in \Bb(0,r)$, we need to find a conservative bound on the radius defined for stability. 
Therefore, a sufficient condition for $R$ to satisfy is the following inequality:
\begin{equation}
    \begin{aligned}
            \frac{L}{\mu}C_1 (R+r)^2 + \frac{2}{\mu}(C_2 + C_3+ C_1 C_4) \leq R^2-2rR,
    \end{aligned}
\end{equation}
and the radius for stability around optimal solution is 
\begin{equation}
    \begin{aligned}
          R =& C_5 r + \sqrt{\frac{1+3\kappa C_1}{(1-\kappa C_1)^2}r^2 + \frac{{2}(C_2 + C_3 +  C_1 C_4)}{\mu - {L}C_1}}.\\
    \end{aligned}
\end{equation}
The proper choices for $b,c$ are $$c = \frac{\eta \beta_1}{1 - \beta_1} ~~~~~~~~~ b = \frac{\eta \sigma^2}{1-\beta_1},$$ which completes our proof. \qed

\subsection{Proof of Lemma \ref{lemma_lya}}\label{sec:lemma}
In this section, we present the detailed proof of Lemma \ref{lemma_lya}.
\begin{proof}

From the $L$-smoothness of function $f$, we can get the following inequality:
\begin{equation}
    f(x^{(k+1)}) - f(x^{(k)}) \leq \frac{L}{2}\norm{x^{(k+1)} - x^{(k)}}^2 - \eta \underbrace{\langle \nabla f(x^{(k)}), (\epsilon I + \hat{V}^{(k+1)})^{-\frac{1}{2}} h^{(k+1)}\rangle}_{-A^{(k)}}.
\end{equation}
For the ease of analysis, we denote the term $-\langle \nabla f(x^{(k)}), (\epsilon I + \hat{V}^{(k+1)})^{-\frac{1}{2}} h^{(k+1)}\rangle$ as  $A^{(k)}$, and next, we consider the upper bound for $A^{(k)}$, which can be decomposed into three parts denoted as $I_1, I_2$ and $I_3$, respectively.
\begin{equation}
    \begin{aligned}
        \mathbb{E}\left[A^{(k)}\big|x^{(k)}\right] =& -\mathbb{E}\left[\langle \nabla f(x^{(k)}), (\epsilon I + \hat{V}^{(k+1)})^{-\frac{1}{2}} h^{(k+1)}\rangle\big|x^{(k)}\right]\\
        =& -\mathbb{E}\left[\langle \nabla f(x^{(k)}), (\epsilon I + \hat{V}^{(k)})^{-\frac{1}{2}} h^{(k+1)}\rangle\big|x^{(k)}\right]\\
        &-\mathbb{E}\left[\langle \nabla f(x^{(k)}), [(\epsilon I + \hat{V}^{(k+1)})^{-\frac{1}{2}}-(\epsilon I + \hat{V}^{(k)})^{-\frac{1}{2}}] h^{(k+1)}\rangle\big|x^{(k)}\right]\\
        =& \underbrace{-\beta_1 \mathbb{E}\left[\langle \nabla f(x^{(k)}), (\epsilon I + \hat{V}^{(k)})^{-\frac{1}{2}} h^{(k)}\rangle\big|x^{(k)}\right]}_{I_1} \\
        &\underbrace{- (1-\beta_1)\mathbb{E}\left[\langle \nabla f(x^{(k)}), (\epsilon I + \hat{V}^{(k)})^{-\frac{1}{2}} \nabla f(x^{(k)},\xi^{(k)})\rangle\big|x^{(k)}\right]}_{I_2}\\
        &\underbrace{-\mathbb{E}\left[\langle \nabla f(x^{(k)}), [(\epsilon I + \hat{V}^{(k+1)})^{-\frac{1}{2}}-(\epsilon I + \hat{V}^{(k)})^{-\frac{1}{2}}] h^{(k+1)}\rangle\big|x^{(k)}\right]}_{I_3}
    \end{aligned}
\end{equation}

Using properties from the functions, we can provide an upper bound for all the three terms above.
\begin{equation}
    \begin{aligned}
        I_1 =& -\beta_1 \mathbb{E}\left[\langle \nabla f(x^{(k)}), (\epsilon I + \hat{V}^{(k)})^{-\frac{1}{2}} h^{(k)}\rangle\big|x^{(k)}\right]\\
         =& -\beta_1 \mathbb{E}\left[\langle \nabla f(x^{(k-1)}), (\epsilon I + \hat{V}^{(k)})^{-\frac{1}{2}} h^{(k)}\rangle\big|x^{(k)}\right] \\
        &+  \beta_1 \mathbb{E}\left[\langle \nabla f(x^{(k-1)}) - \nabla f(x^{(k)}), (\epsilon I + \hat{V}^{(k)})^{-\frac{1}{2}} h^{(k)}\rangle\big|x^{(k)}\right]\\
        \leq & \beta_1 A^{(k-1)} +\frac{\beta_1 L}{\eta}\norm{x^{(k)} - x^{(k-1)}}^2
    \end{aligned}
\end{equation}

\begin{equation}
    \begin{aligned}
        I_2 =& - (1-\beta_1)\mathbb{E}\left[\langle \nabla f(x^{(k)}), (\epsilon I + \hat{V}^{(k)})^{-\frac{1}{2}} \nabla f(x^{(k)},\xi^{(k)})\rangle\big|x^{(k)}\right]\\
        =&- (1-\beta_1)\langle \nabla f(x^{(k)}), (\epsilon I + \hat{V}^{(k)})^{-\frac{1}{2}} \nabla f(x^{(k)})\rangle\\
        =&- (1-\beta_1) \norm{\nabla f(x^{(k)})}^2_{(\epsilon I + \hat{V}^{(k)})^{-\frac{1}{2}}}
    \end{aligned}
\end{equation}

\begin{equation}
    \begin{aligned}
        I_3 = & -\mathbb{E}\left[\langle \nabla f(x^{(k)}), [(\epsilon I + \hat{V}^{(k+1)})^{-\frac{1}{2}}-(\epsilon I + \hat{V}^{(k)})^{-\frac{1}{2}}] h^{(k+1)}\rangle\big|x^{(k)}\right]\\
        \leq & \mathbb{E}\left[ \norm{\nabla f(x^{(k)})}\norm{h^{(k+1)}}\norm{(\epsilon I + \hat{V}^{(k+1)})^{-\frac{1}{2}}-(\epsilon I + \hat{V}^{(k)})^{-\frac{1}{2}}}  \big|x^{(k)}\right]\\
        \leq & { \sigma^2 \mathbb{E}\left[\sum_{i=1}^d [(\epsilon + [\hat{v}^{(k)}]_i)^{-\frac{1}{2}} - (\epsilon + [\hat{v}^{(k+1)}]_i)^{-\frac{1}{2}}] \Big|x^{(k)}\right]}
    \end{aligned}
\end{equation}
Then, we can write the relationship of Lyapunov function between two consecutive iterations.
\begin{equation}
\resizebox{\hsize}{!}{$
    \begin{aligned}
    \mathbb{E}\left[\mathcal{L}^{(k+1)}\big|x^{(k)}\right] & -  \mathcal{L}^{(k)}  =  \mathbb{E}\left[f(x^{(k+1)}) - c \langle \nabla f(x^{(k)}), (\epsilon I + \hat{V}^{(k+1)})^{-\frac{1}{2}} h^{(k+1)}\rangle  + b \sum_{i=1}^d (\epsilon + [\hat{v}^{(k+1)}]_i)^{-\frac{1}{2}}- f(x^{(k)})\Big|x^{(k)}\right]\\
    & + c \langle \nabla f(x^{(k-1)}), (\epsilon I + \hat{V}^{(k)})^{-\frac{1}{2}} h^{(k)}\rangle  - b \sum_{i=1}^d (\epsilon + [\hat{v}^{(k)}]_i)^{-\frac{1}{2}}\\
    =&\ex{f(x^{(k+1)}) - f(x^{(k)})\Big|x^{(k)}} + c \mathbb{E}\left[A^{(k)}\Big|x^{(k)}\right] - c A^{(k-1)}  + b \mathbb{E}\left[\sum_{i=1}^d (\epsilon + [\hat{v}^{(k+1)}]_i)^{-\frac{1}{2}}- \sum_{i=1}^d (\epsilon + [\hat{v}^{(k)}]_i)^{-\frac{1}{2}}\Big|x^{(k)}\right]\\
    \leq & \mathbb{E}\left[\frac{L}{2}\norm{x^{(k+1)} - x^{(k)}}^2 \Big|x^{(k)} \right]+ (\eta+c) \mathbb{E}\left[A^{(k)}\Big|x^{(k)}\right]  - c A^{(k-1)} + b \mathbb{E}\left[\sum_{i=1}^d (\epsilon + [\hat{v}^{(k+1)}]_i)^{-\frac{1}{2}}- \sum_{i=1}^d (\epsilon + [\hat{v}^{(k)}]_i)^{-\frac{1}{2}}\Big|x^{(k)}\right]\\
    \leq 
    & \mathbb{E}\left[\frac{L}{2}\norm{x^{(k+1)} - x^{(k)}}^2 \Big|x^{(k)} \right]  - c A^{(k-1)}  + b \mathbb{E}\left[\sum_{i=1}^d (\epsilon + [\hat{v}^{(k+1)}]_i)^{-\frac{1}{2}}- \sum_{i=1}^d (\epsilon + [\hat{v}^{(k)}]_i)^{-\frac{1}{2}}\Big|x^{(k)}\right]\\
    &+ (\eta + c)\bigg(\beta_1 A^{(k-1)} +\frac{\beta_1 L}{\eta}\norm{x^{(k)} - x^{(k-1)}}^2  - (1-\beta_1) \norm{\nabla f(x^{(k)})}^2_{(\epsilon I + \hat{V}^{(k)})^{-\frac{1}{2}}} \\
    &+ \sigma^2 \mathbb{E}\left[\sum_{i=1}^d [(\epsilon + [\hat{v}^{(k)}]_i)^{-\frac{1}{2}} - (\epsilon + [\hat{v}^{(k+1)}]_i)^{-\frac{1}{2}}] \Big|x^{(k)}\right] \bigg )\\
    \leq 
    & \frac{L}{2}\frac{\eta^2 d}{(1-\beta_2)(1 - \beta_1^2/\beta_2)} - c A^{(k-1)}  + b \mathbb{E}\left[\sum_{i=1}^d (\epsilon + [\hat{v}^{(k+1)}]_i)^{-\frac{1}{2}}- \sum_{i=1}^d (\epsilon + [\hat{v}^{(k)}]_i)^{-\frac{1}{2}}\Big|x^{(k)}\right]\\
    &+ (\eta + c)\bigg(\beta_1 A^{(k-1)} +\frac{\beta_1 L}{\eta}\frac{\eta^2 d}{(1-\beta_2)(1 - \beta_1^2/\beta_2)}  - (1-\beta_1) \norm{\nabla f(x^{(k)})}^2_{(\epsilon I + \hat{V}^{(k)})^{-\frac{1}{2}}} \\
    &+ \sigma^2 \mathbb{E}\left[\sum_{i=1}^d [(\epsilon + [\hat{v}^{(k)}]_i)^{-\frac{1}{2}} - (\epsilon + [\hat{v}^{(k+1)}]_i)^{-\frac{1}{2}}] \Big|x^{(k)}\right] \bigg ).
    \end{aligned}
    $}
\end{equation}
The last inequality is based on Lemma 8 in \cite{chen2021cada}. Now, if we set $b = \sigma^2(\eta + c)$ and$ (\eta + c)\beta_1 = c$, i.e., $$c = \frac{\eta \beta_1}{1 - \beta_1},~~~~~~~ b = \frac{\eta \sigma^2}{1-\beta_1},$$  the equation above can be greatly simplified.
\begin{equation}
    \begin{aligned}
    \mathbb{E}\left[\mathcal{L}^{(k+1)}\big|x^{(k)}\right]   \leq & \mathcal{L}^{(k)} + (\frac{L}{2}+ (\eta + c)\frac{\beta_1 L}{\eta})\frac{\eta^2 d}{(1-\beta_2)(1 - \beta_1^2/\beta_2)}  - (\eta+ c) (1-\beta_1) \norm{\nabla f(x^{(k)})}^2_{(\epsilon I + \hat{V}^{(k)})^{-\frac{1}{2}}} \\
    \leq &\mathcal{L}^{(k)} -(\eta + c) (1-\beta_1) \norm{\nabla f(x^{(k)})}^2_{(\epsilon I + \hat{V}^{(k)})^{-\frac{1}{2}}}  + (\frac{L}{2}+ \frac{ L\beta_1}{1-\beta_1})\frac{\eta^2 d}{(1-\beta_2)(1 - \beta_1^2/\beta_2)}\\
    \leq & \mathcal{L}^{(k)}-\eta  (\epsilon + {\sigma^2})^{-\frac{1}{2}}\norm{\nabla f(x^{(k)})}^2  + (\frac{L}{2}+ \frac{ L\beta_1}{1-\beta_1})\frac{\eta^2 d}{(1-\beta_2)(1 - \beta_1^2/\beta_2)}\\
    \leq &\mathcal{L}^{(k)} -2 \mu\eta (\epsilon + \sigma^2)^{-\frac{1}{2}}(f(x^{(k)}) - f(x^\star))  + (\frac{L}{2}+ \frac{ L\beta_1}{1-\beta_1})\frac{\eta^2 d}{(1-\beta_2)(1 - \beta_1^2/\beta_2)}\\
    \leq & \mathcal{L}^{(k)}-2 \mu\eta  (\epsilon + \sigma^2)^{-\frac{1}{2}}\Big(\mathcal{L}^{(k)} - b \sum_{i=1}^d (\epsilon + [\hat{v}^{(k)}]_i)^{-\frac{1}{2}} + c \langle \nabla f(x^{(k-1)}), (\epsilon I + \hat{V}^{(k)})^{-\frac{1}{2}} h^{(k)}\rangle \Big) \\
    & + (\frac{L}{2}+ \frac{ L\beta_1}{1-\beta_1})\frac{\eta^2 d}{(1-\beta_2)(1 - \beta_1^2/\beta_2)}\\
    =& (1 - 2 \mu\eta (\epsilon + \sigma^2)^{-\frac{1}{2}}) \mathcal{L}^{(k)}+ (\frac{L}{2}+ \frac{ L\beta_1}{1-\beta_1})\frac{\eta^2 d}{(1-\beta_2)(1 - \beta_1^2/\beta_2)}\\
    &  -2 \mu\eta  (\epsilon + \sigma^2)^{-\frac{1}{2}} \Big(- b \sum_{i=1}^d (\epsilon + [\hat{v}^{(k)}]_i)^{-\frac{1}{2}} +c \langle \nabla f(x^{(k-1)}), (\epsilon I + \hat{V}^{(k)})^{-\frac{1}{2}} h^{(k)}\rangle  \Big)\\
    \leq & (1 - 2 \mu\eta  (\epsilon + \sigma^2)^{-\frac{1}{2}}) \mathcal{L}^{(k)}+ (\frac{L}{2}+ \frac{ L\beta_1}{1-\beta_1})\frac{\eta^2 d}{(1-\beta_2)(1 - \beta_1^2/\beta_2)}\\
    &  +2 \mu\eta^2  (\epsilon + \sigma^2)^{-\frac{1}{2}}  \sigma^2 d \epsilon^{-\frac{1}{2}}\left(\frac{1+\beta_1}{1-\beta_1}\right)  \\
    =& C_1 \mathcal{L}^{(k)} + C_2
    \end{aligned}
\end{equation}
\end{proof}

        

\section{Real-World Data Experiments}

\begin{figure}[h!]
\begin{subfigure}{0.235\textwidth}
  \centering
  \includegraphics[width=1\linewidth]{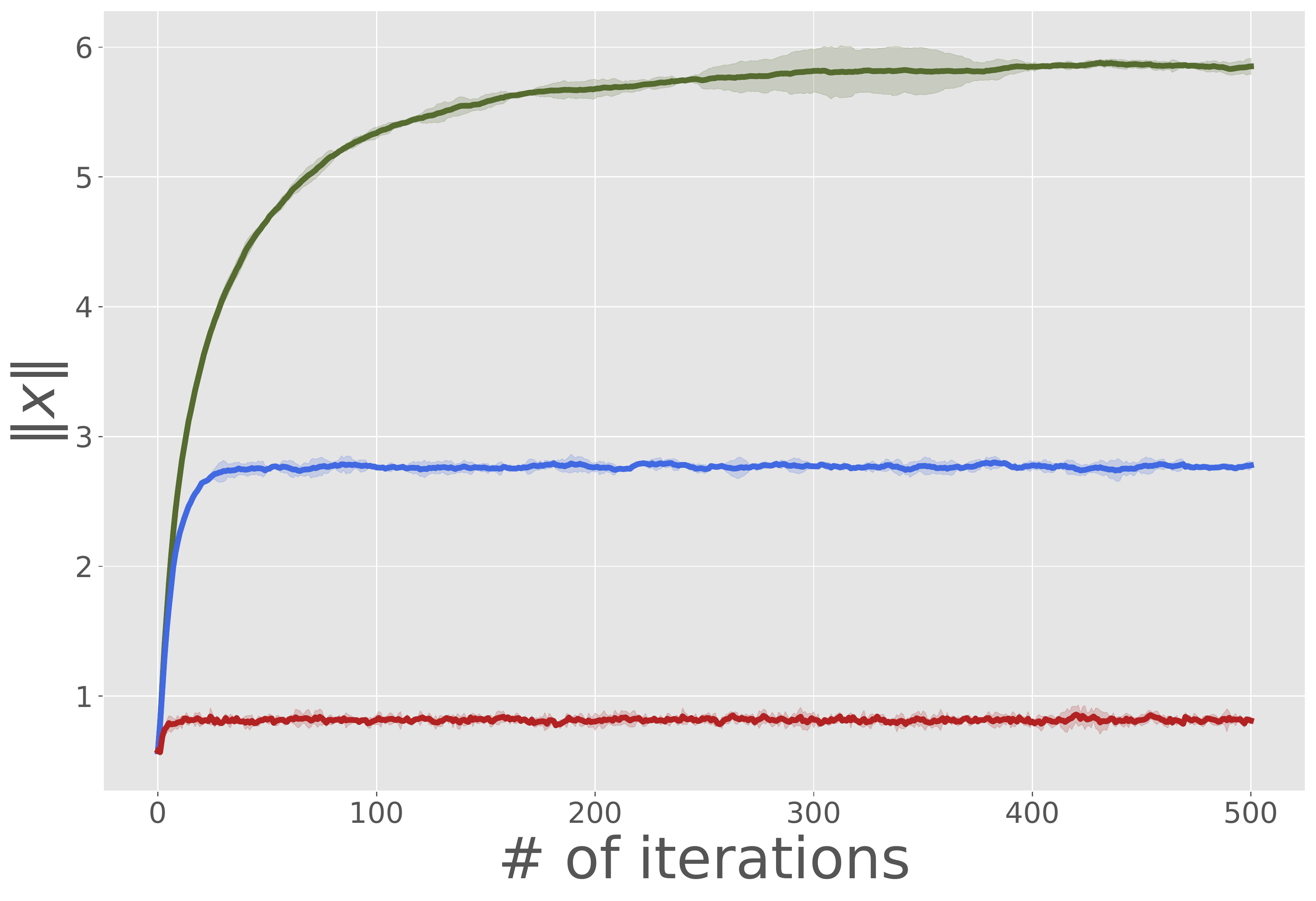}
  \caption{Iterate norm}
\end{subfigure}
\hfill
\begin{subfigure}{0.235\textwidth}
  \centering
  \includegraphics[width=1\linewidth]{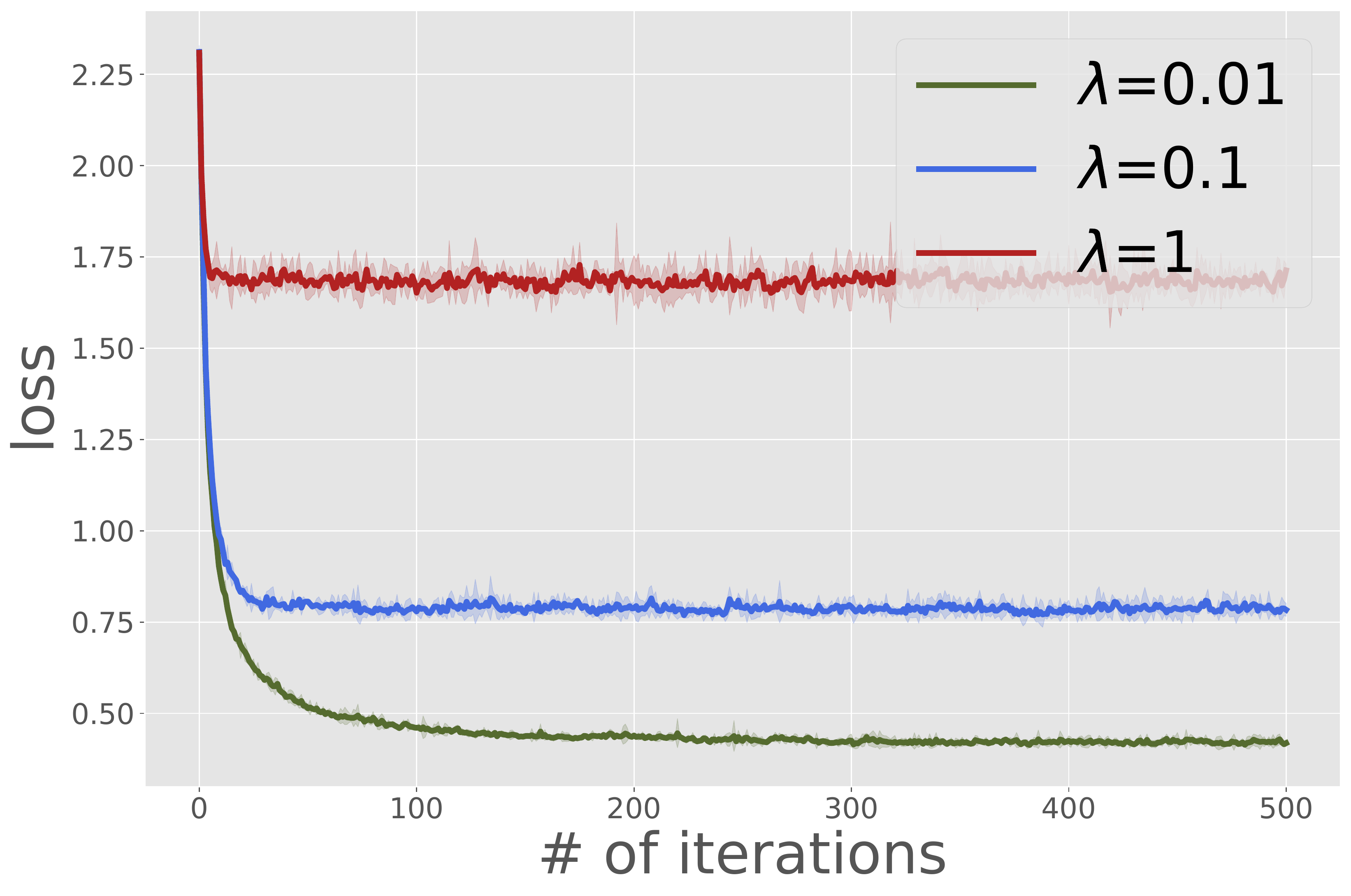}
  \caption{Loss}
\end{subfigure}
\hfill
\begin{subfigure}{0.235\textwidth}
  \centering
  \includegraphics[width=1\linewidth]{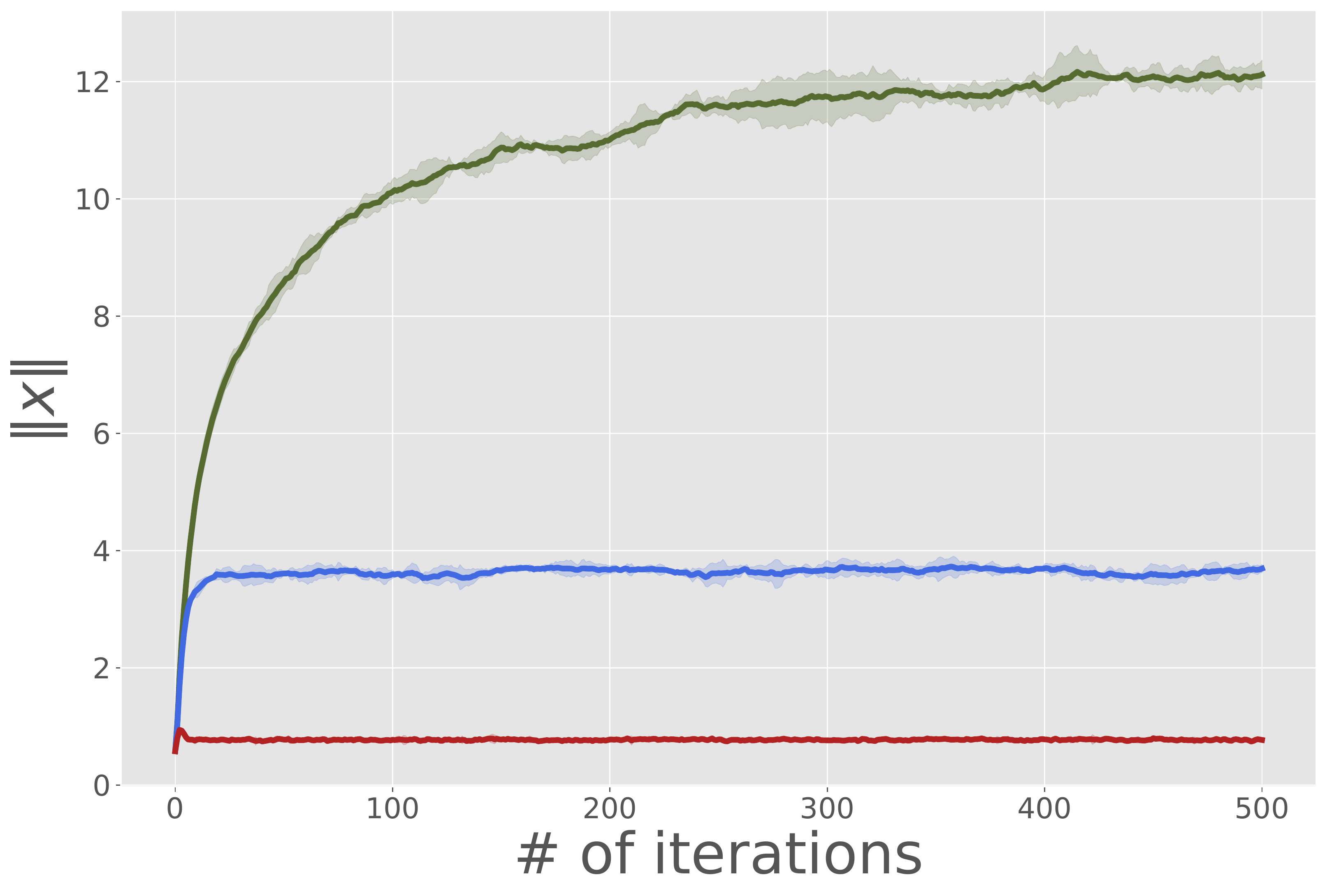}
  \caption{Iterate norm}
\end{subfigure}
\hfill
\begin{subfigure}{0.235\textwidth}
  \centering
  \includegraphics[width=1\linewidth]{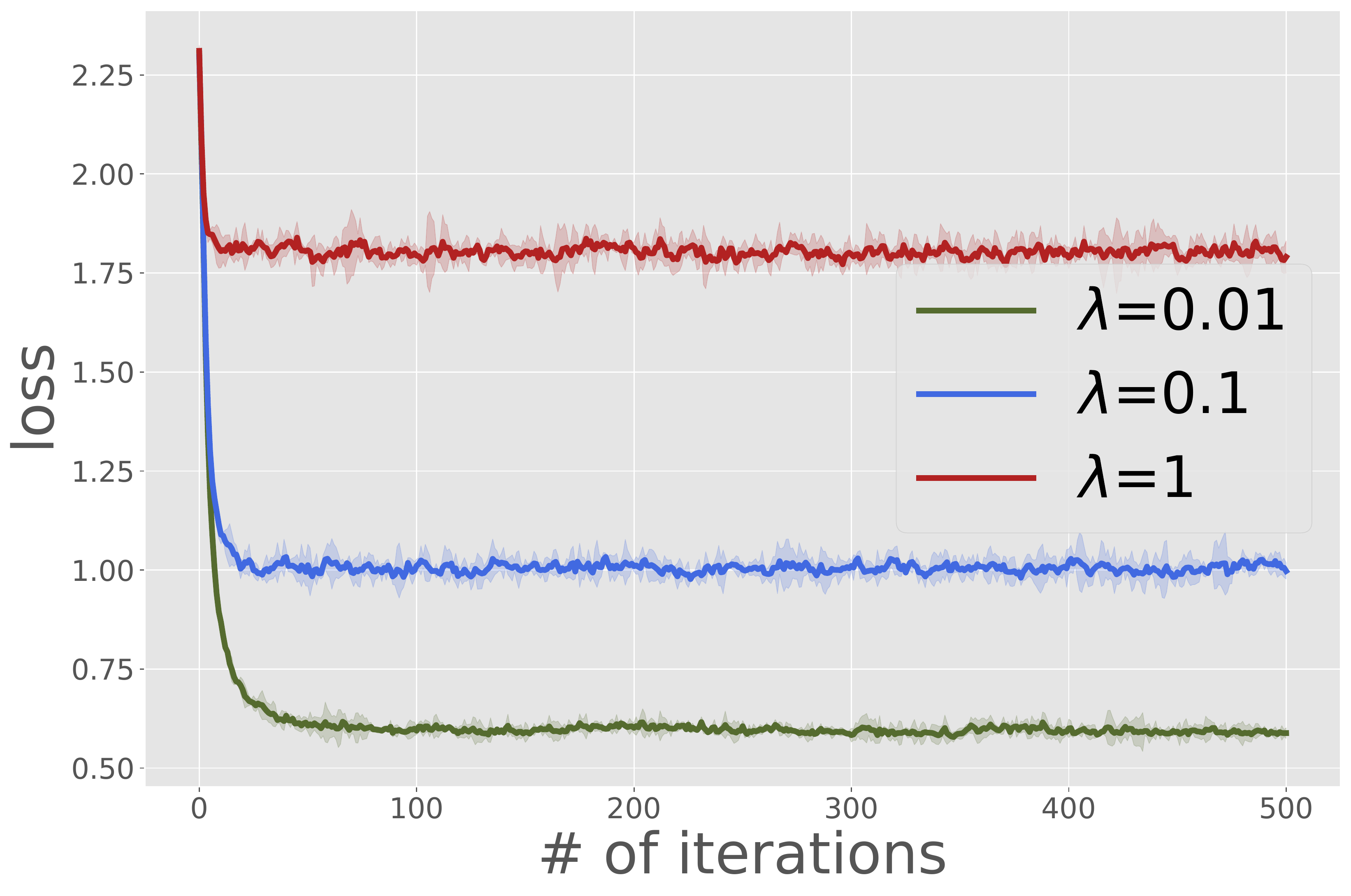}
  \caption{Loss}
\end{subfigure}
\caption{Comparison of iterate norms and loss for different $\lambda$ for local SGD (left) and local Adam (right).} 
\label{fig:mnist-adam-lambda}
\vspace{-0.2cm}
\end{figure}

\begin{figure}[t]
\begin{subfigure}{0.235\textwidth}
  \centering
  \includegraphics[width=1\linewidth]{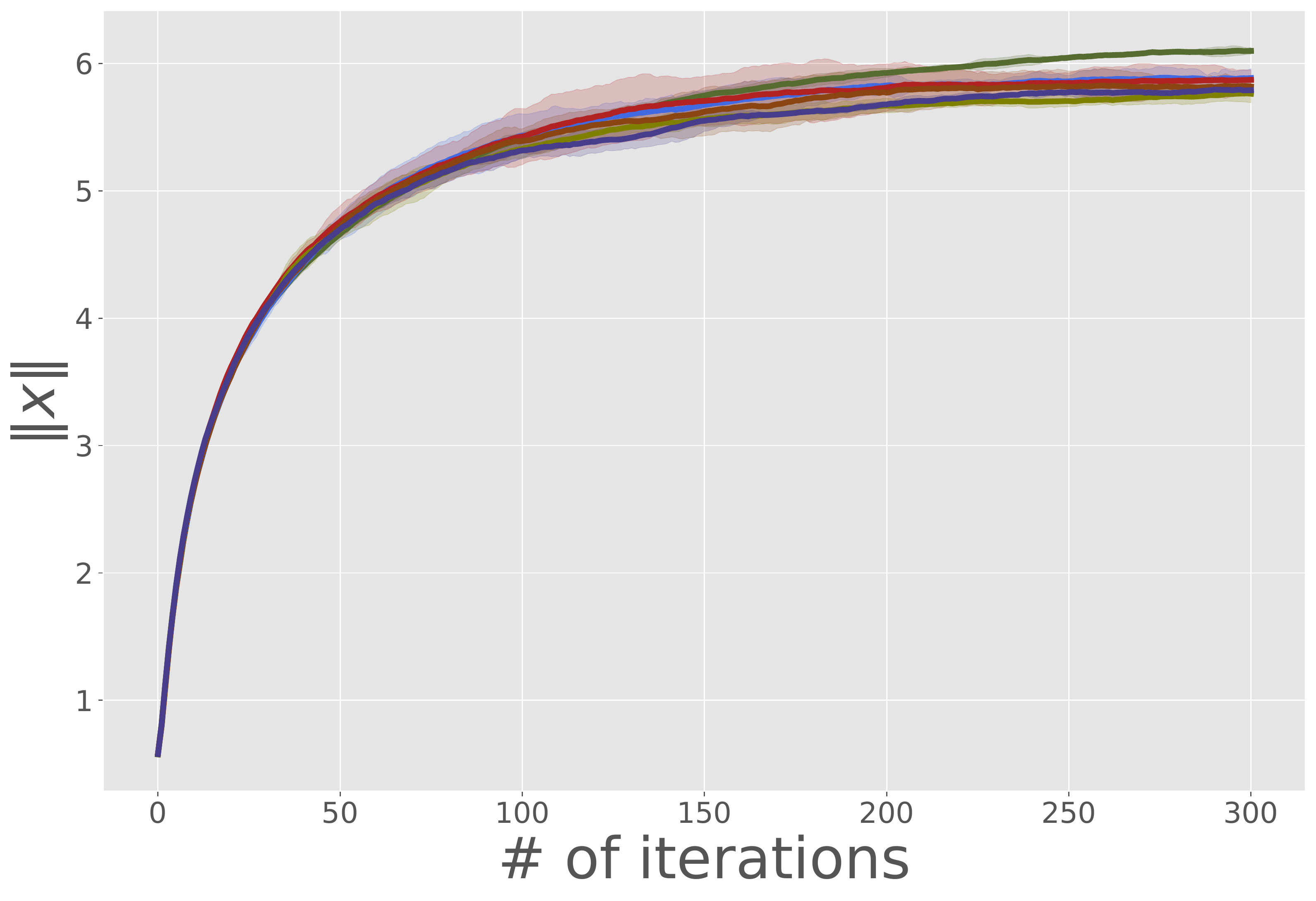}
  \caption{Iterate norm}
\end{subfigure}
\hfill
\begin{subfigure}{0.235\textwidth}
  \centering
  \includegraphics[width=1\linewidth]{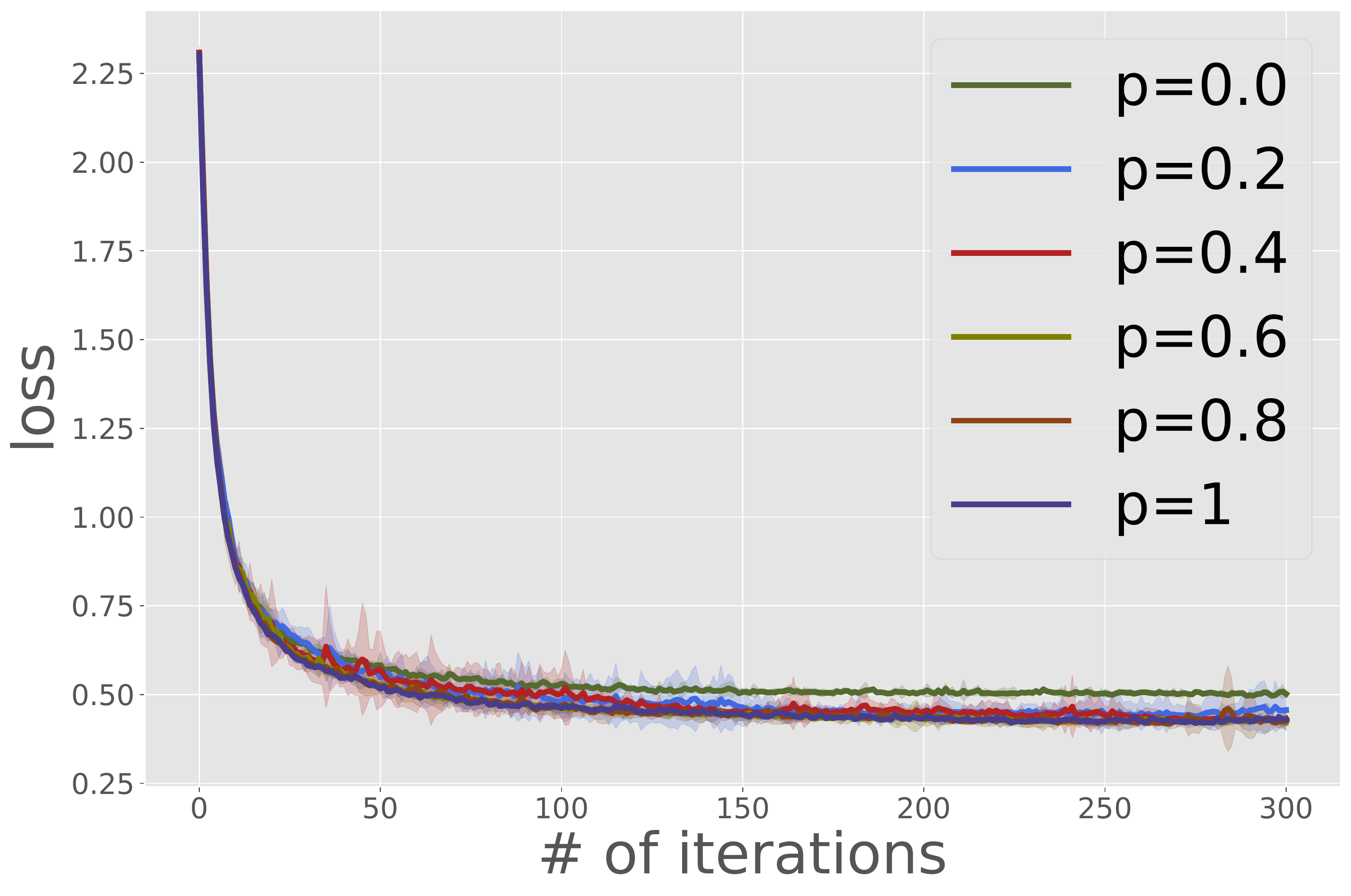}
  \caption{Loss}
\end{subfigure}
\hfill
\begin{subfigure}{0.235\textwidth}
  \centering
  \includegraphics[width=1\linewidth]{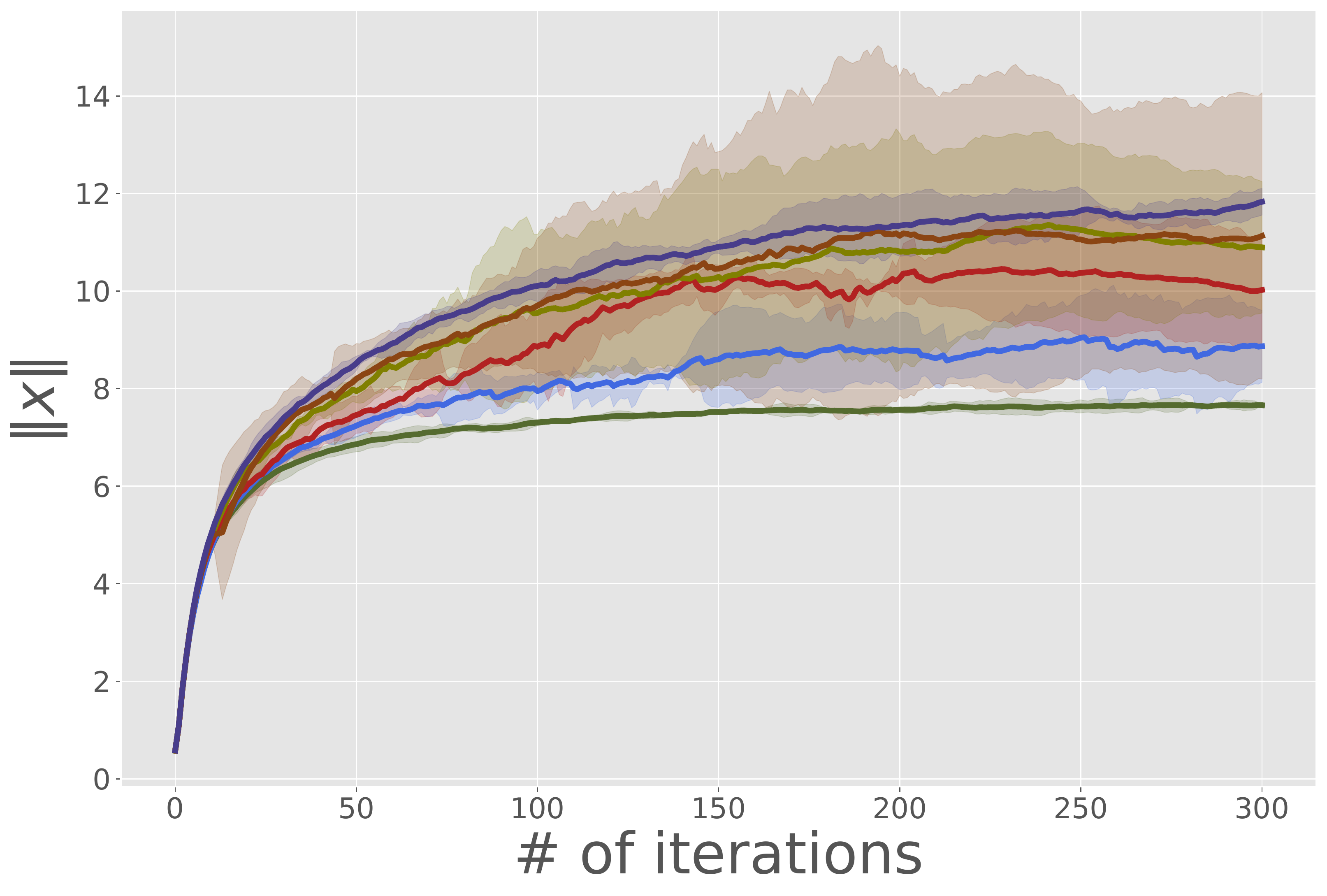}
  \caption{Iterate norm}
\end{subfigure}
\hfill
\begin{subfigure}{0.235\textwidth}
  \centering
  \includegraphics[width=1\linewidth]{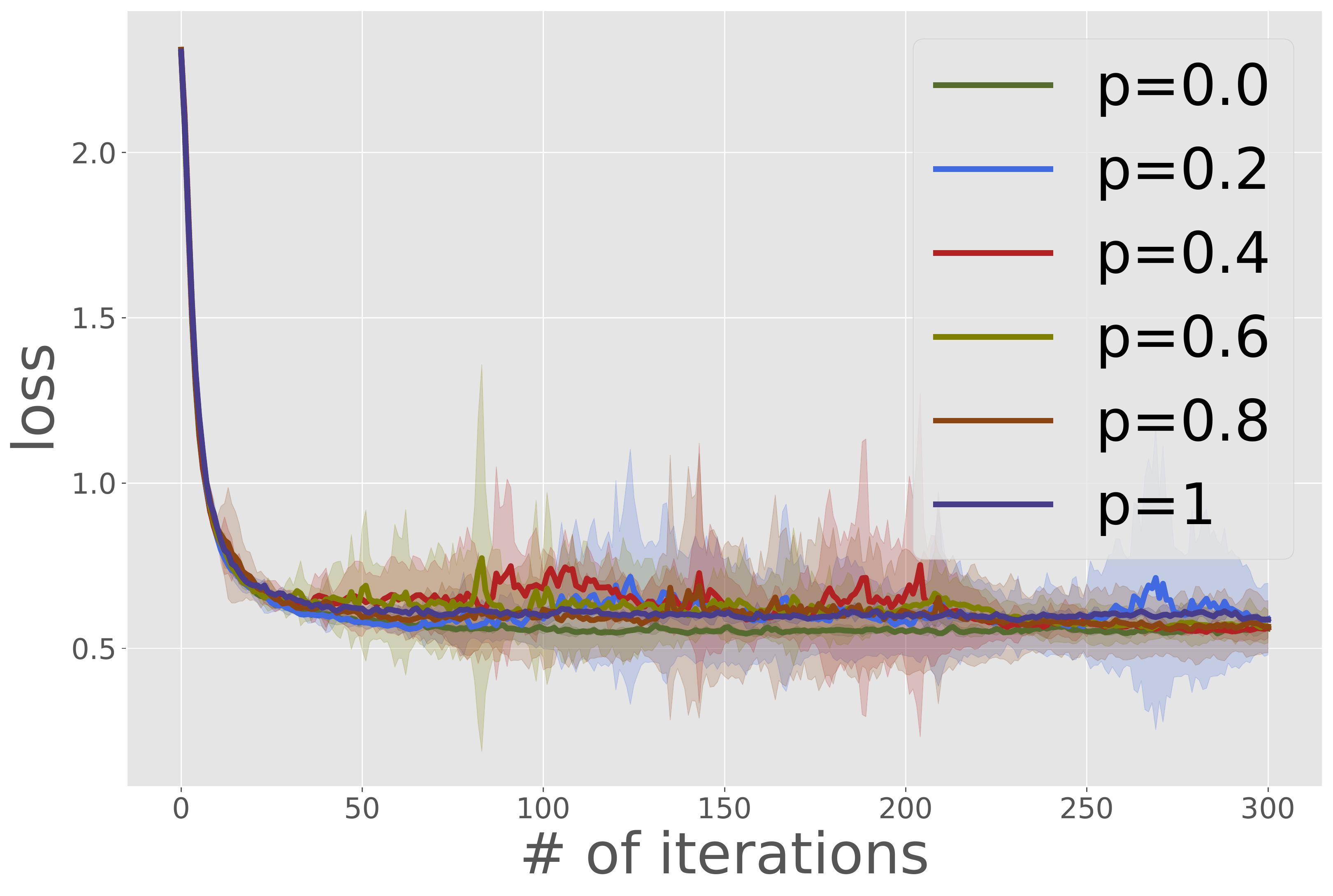}
  \caption{Loss}
\end{subfigure}
\caption{Comparison of iterate norms and loss for different $p$ for local SGD (left) and local Adam (right).} 
\label{fig:mnist-adam-p}
\vspace{-0.2cm}
\end{figure}

\subsection{MNIST dataset}

We consider a similar simulation setting as in Section \ref{sec:synth}, except with real world data distributed among clients. Specifically, we use the  MNIST \cite{deng2012mnist} dataset, which consists of 60,000 examples of handwritten digits. The data set consists of $10$ label classes with each data of dimension $d=784$. We distribute the data among the clients independently and identically such that each client has $m=100$ data. The system consists of a pool of $N=600$ clients, and in each simulation the current number of clients initializes at $N_0=10$. The local loss function at each client is the $\ell_2$ regularized cross-entropy. In this setting we analyze the global loss in addition to the norm of the global iterate, where the global loss is the average of the local loss for a given global iterate. For  MNIST data set, the local optimization parameters for local SGD are set as $\eta=0.05$, and for local Adam as $\beta_1=0.9$, $\beta_2=0.999$, $\epsilon=10^{-8}$ and $\eta=0.005$.

Fig. 
\ref{fig:mnist-adam-lambda} shows how the global iterate norm  and the global loss change with the   parameter $\lambda$ for local SGD and local Adam, for $p=1$. Similar to the synthetic   setting, in both local SGD and local Adam, the stable value of global iterate norm is higher when $\lambda$ is smaller.

Fig. 
\ref{fig:mnist-adam-p} shows how the global iterate and loss behave with changing $p$ for local SGD and local Adam. The global loss for both algorithms for different $p$ differs only slightly, while the variance of the results being high for larger values of $p$. In both cases of local Adam and local SGD, the global iterate maintains stability, for all possible $p$ considered. 

\section{Additional Experiments}

In addition to the convex logistic loss, we conduct experiments with convolutional neural networks (CNNs) as client models, in the same open FL system setting described in Section 4. Due to the choice of client models, the landscape of the local objective functions  in this case is not guaranteed to be convex. The CNN model at each client consists of two convolutional layers, each followed by a ReLU activation layer and a max pooling layer. 

We distribute the data among the clients independently and identically such that each client has $m=100$ data. The system consists of a pool of $N=600$ clients, and in each simulation the current number of clients initializes at $N_0=10$. The local loss function at each client is the $\ell_2$ regularized cross-entropy. Similar to the experiments described in Section 4, we investigate the effect of the regularization parameter $\lambda$ of the local loss function and the probability $p$ of clients joining and/or leaving the system. We provide results for local Adam, since the behavior of local SGD and local Adam are similar. The local optimization parameters of clients are set as $\beta_1=0.9$, $\beta_2=0.999$, $\epsilon=10^{-8}$ and $\eta=0.005$.


\begin{figure}[h!]
\centering
\begin{subfigure}{0.45\textwidth}
  \centering
  \includegraphics[width=1\linewidth]{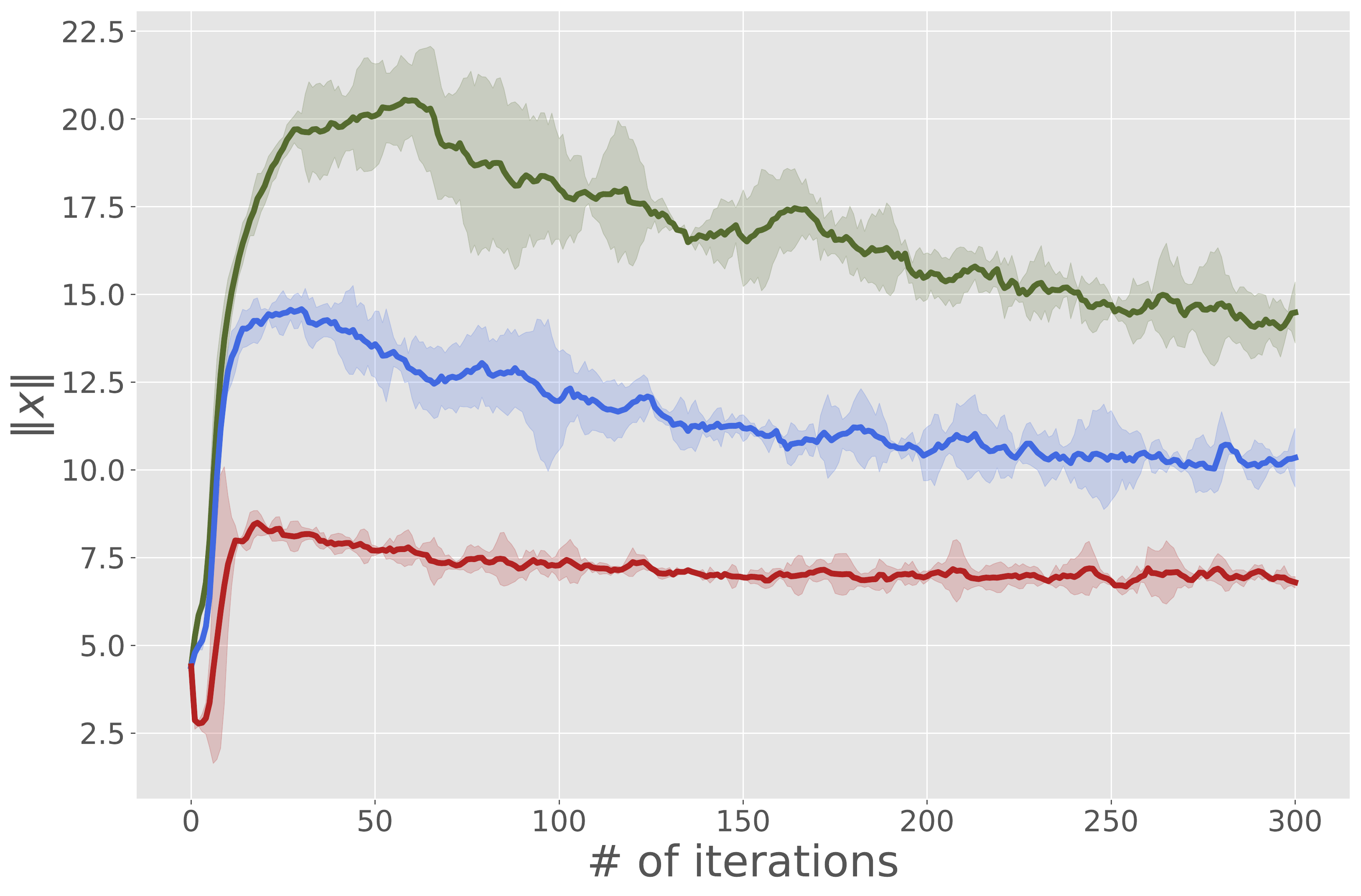}
  \caption{Iterate norm}
\end{subfigure}
\begin{subfigure}{0.45\textwidth}
  \centering
  \includegraphics[width=1\linewidth]{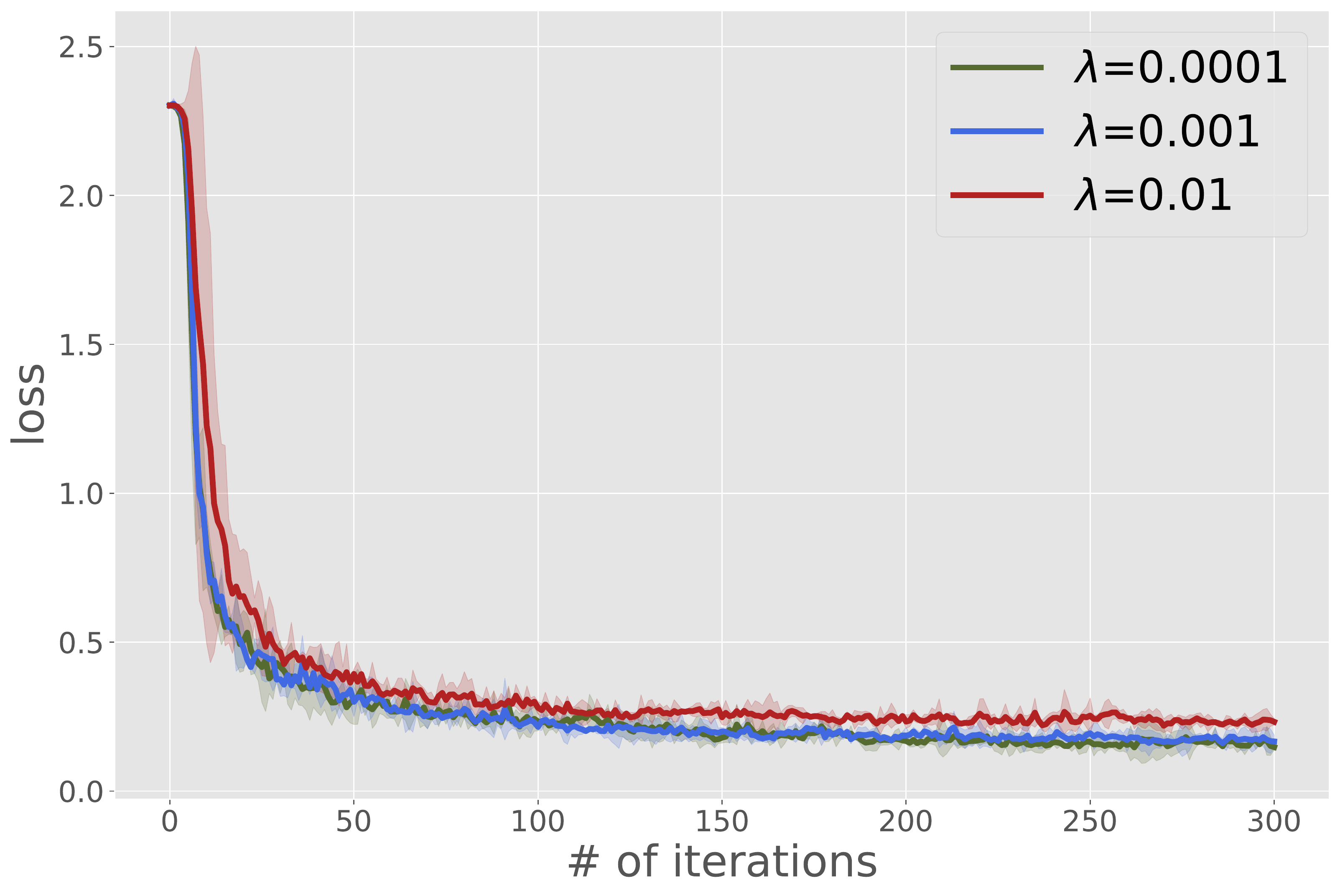}
  \caption{Loss}
\end{subfigure}
\caption{Comparison of iterate norms and loss for different $\lambda$ for local Adam.} 
\label{fig:mnist-cnn-adam-lambda}
\vspace{-0.2cm}
\end{figure}


\begin{figure}[h!]
\centering
\begin{subfigure}{0.45\textwidth}
  \centering
  \includegraphics[width=1\linewidth]{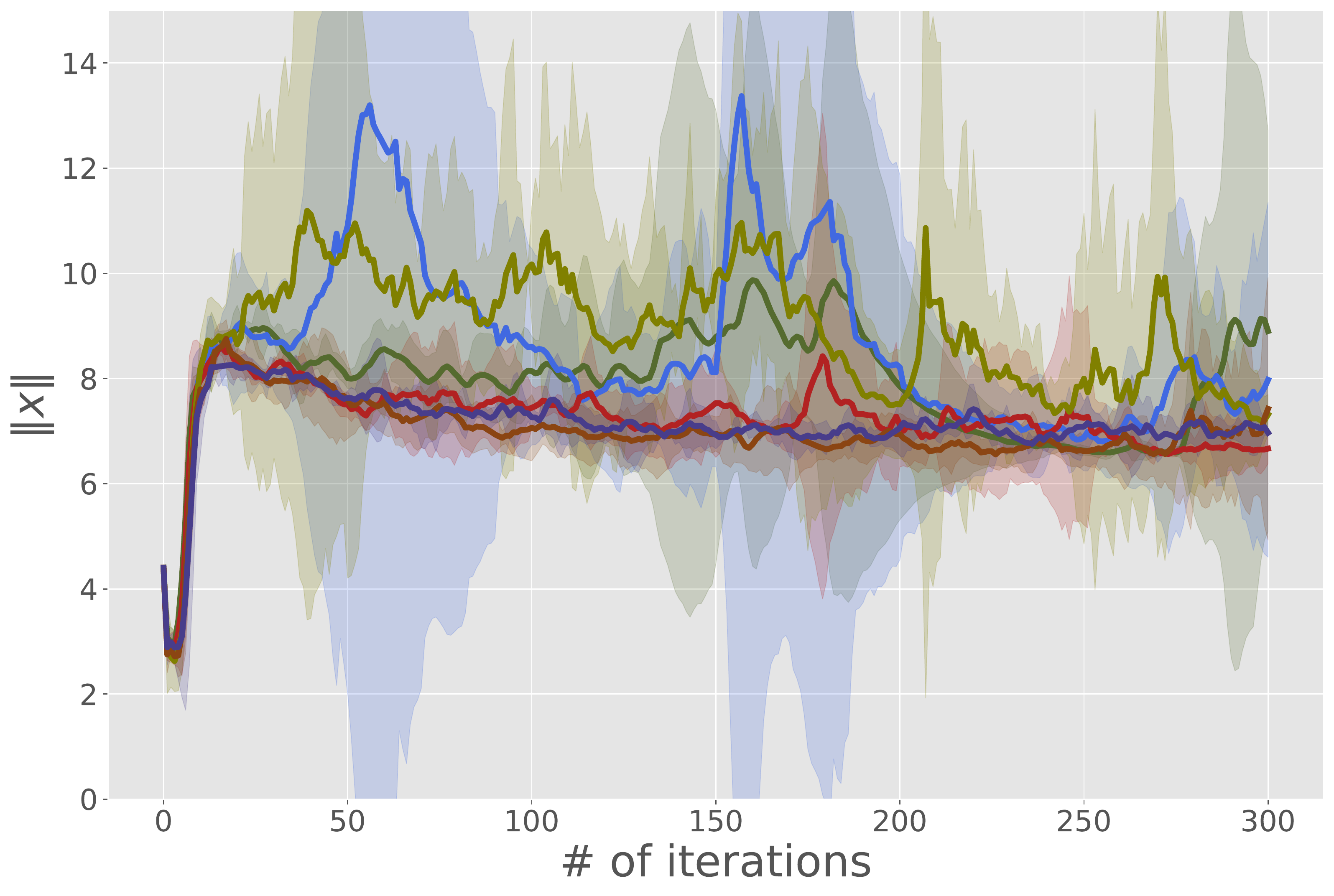}
  \caption{Iterate norm}
\end{subfigure}
\begin{subfigure}{0.45\textwidth}
  \centering
  \includegraphics[width=1\linewidth]{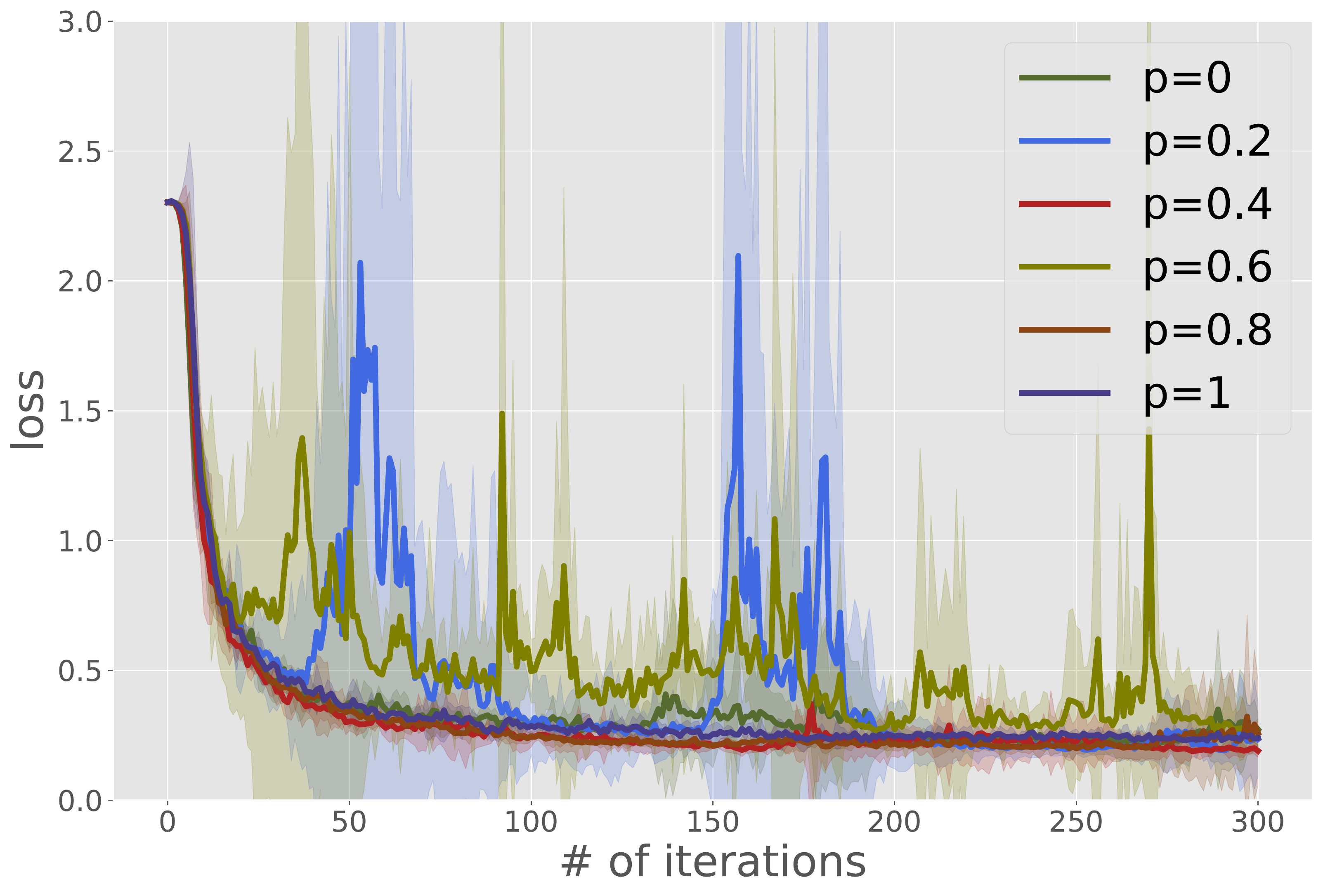}
  \caption{Loss}
\end{subfigure}
\caption{Comparison of iterate norms and loss for different $p$ for local Adam.} 
\label{fig:mnist-cnn-adam-p}
\vspace{-0.2cm}
\end{figure}

Fig. \ref{fig:mnist-cnn-adam-lambda} shows how the global iterate norm  and the global loss change with the regularization parameter $\lambda$ for local Adam, for $p=1$. It can be seen that the global iterate norm converges to a stable value after some iterations for the tested $\lambda$ values, even though the local models are not guaranteed to be convex. As expected, the stable value of global iterate norm is higher when $\lambda$ is smaller. 

 Fig. \ref{fig:mnist-cnn-adam-p} shows how the global iterate norm and the global loss behave with changing $p$ for local Adam. The global loss for different $p$ differs only slightly but shows high variance in some cases due to the dynamic nature of the system. Accordingly, the global iterate seems to achieve stability after some number of iterations for all tested $p$. These empirical results suggest that non-convex models like CNNs show stability in open FL systems.

%% file: root.bbl
\begin{thebibliography}{10}
\providecommand{\url}[1]{#1}
\csname url@samestyle\endcsname
\providecommand{\newblock}{\relax}
\providecommand{\bibinfo}[2]{#2}
\providecommand{\BIBentrySTDinterwordspacing}{\spaceskip=0pt\relax}
\providecommand{\BIBentryALTinterwordstretchfactor}{4}
\providecommand{\BIBentryALTinterwordspacing}{\spaceskip=\fontdimen2\font plus
\BIBentryALTinterwordstretchfactor\fontdimen3\font minus
  \fontdimen4\font\relax}
\providecommand{\BIBforeignlanguage}[2]{{%
\expandafter\ifx\csname l@#1\endcsname\relax
\typeout{** WARNING: IEEEtran.bst: No hyphenation pattern has been}%
\typeout{** loaded for the language `#1'. Using the pattern for}%
\typeout{** the default language instead.}%
\else
\language=\csname l@#1\endcsname
\fi
#2}}
\providecommand{\BIBdecl}{\relax}
\BIBdecl

\bibitem{mcmahan2017communication}
B.~McMahan, E.~Moore, D.~Ramage, S.~Hampson, and B.~A. y~Arcas,
  ``Communication-efficient learning of deep networks from decentralized
  data,'' in \emph{Artificial Intelligence and Statistics (AISTATS)}.\hskip 1em
  plus 0.5em minus 0.4em\relax PMLR, 2017, pp. 1273--1282.

\bibitem{woodworth2020local}
B.~Woodworth, K.~K. Patel, S.~Stich, Z.~Dai, B.~Bullins, B.~Mcmahan, O.~Shamir,
  and N.~Srebro, ``Is local sgd better than minibatch sgd?'' in
  \emph{International Conference on Machine Learning (ICML)}.\hskip 1em plus
  0.5em minus 0.4em\relax PMLR, 2020, pp. 10\,334--10\,343.

\bibitem{li2020federated}
T.~Li, A.~K. Sahu, A.~Talwalkar, and V.~Smith, ``Federated learning:
  Challenges, methods, and future directions,'' \emph{IEEE Signal Processing
  Magazine}, vol.~37, no.~3, pp. 50--60, 2020.

\bibitem{konevcny2016federated}
J.~Kone{\v{c}}n{\`y}, H.~B. McMahan, F.~X. Yu, P.~Richt{\'a}rik, A.~T. Suresh,
  and D.~Bacon, ``Federated learning: Strategies for improving communication
  efficiency,'' \emph{Advances in Neural Information Processing Systems;
  Private Multi-Party Machine Learning Workshop}, 2016.

\bibitem{li2020federatedoptimization}
T.~Li, A.~K. Sahu, M.~Zaheer, M.~Sanjabi, A.~Talwalkar, and V.~Smith,
  ``Federated optimization in heterogeneous networks,'' \emph{Proceedings of
  Machine learning and Systems}, vol.~2, pp. 429--450, 2020.

\bibitem{wang2020tackling}
J.~Wang, Q.~Liu, H.~Liang, G.~Joshi, and H.~V. Poor, ``Tackling the objective
  inconsistency problem in heterogeneous federated optimization,''
  \emph{Advances in Neural Information Processing Systems (NeurIPS)}, vol.~33,
  pp. 7611--7623, 2020.

\bibitem{bhowmick2018protection}
A.~Bhowmick, J.~Duchi, J.~Freudiger, G.~Kapoor, and R.~Rogers, ``Protection
  against reconstruction and its applications in private federated learning,''
  \emph{arXiv preprint arXiv:1812.00984}, 2018.

\bibitem{sahu2018convergence}
A.~K. Sahu, T.~Li, M.~Sanjabi, M.~Zaheer, A.~Talwalkar, and V.~Smith, ``On the
  convergence of federated optimization in heterogeneous networks,''
  \emph{arXiv preprint arXiv:1812.06127}, vol.~3, p.~3, 2018.

\bibitem{reisizadeh2020fedpaq}
A.~Reisizadeh, A.~Mokhtari, H.~Hassani, A.~Jadbabaie, and R.~Pedarsani,
  ``Fedpaq: A communication-efficient federated learning method with periodic
  averaging and quantization,'' in \emph{International Conference on Artificial
  Intelligence and Statistics (AISTATS)}.\hskip 1em plus 0.5em minus
  0.4em\relax PMLR, 2020, pp. 2021--2031.

\bibitem{karimireddy2020scaffold}
S.~P. Karimireddy, S.~Kale, M.~Mohri, S.~Reddi, S.~Stich, and A.~T. Suresh,
  ``Scaffold: Stochastic controlled averaging for federated learning,'' in
  \emph{International Conference on Machine Learning (ICML)}.\hskip 1em plus
  0.5em minus 0.4em\relax PMLR, 2020, pp. 5132--5143.

\bibitem{zinkevich2010parallelized}
M.~Zinkevich, M.~Weimer, A.~J. Smola, and L.~Li, ``Parallelized stochastic
  gradient descent.'' in \emph{Neural Information Processing Systems (NIPS)},
  vol.~4, no.~1, 2010, p.~4.

\bibitem{stich2018local}
S.~U. Stich, ``Local sgd converges fast and communicates little,'' in
  \emph{International Conference on Learning Representations (ICLR)}, no. CONF,
  2019.

\bibitem{stich2019error}
S.~U. Stich and S.~P. Karimireddy, ``The error-feedback framework: Better rates
  for sgd with delayed gradients and compressed updates,'' \emph{The Journal of
  Machine Learning Research (JMLR)}, vol.~21, no.~1, pp. 9613--9648, 2020.

\bibitem{khaled2020tighter}
A.~Khaled, K.~Mishchenko, and P.~Richt{\'a}rik, ``Tighter theory for local sgd
  on identical and heterogeneous data,'' in \emph{International Conference on
  Artificial Intelligence and Statistics (AISTATS)}.\hskip 1em plus 0.5em minus
  0.4em\relax PMLR, 2020, pp. 4519--4529.

\bibitem{mitra2021achieving}
A.~Mitra, R.~Jaafar, G.~J. Pappas, and H.~Hassani, ``Linear convergence in
  federated learning: Tackling client heterogeneity and sparse gradients,''
  \emph{Advances in Neural Information Processing Systems (NeurIPS)}, vol.~34,
  pp. 14\,606--14\,619, 2021.

\bibitem{yu2019linear}
H.~Yu, R.~Jin, and S.~Yang, ``On the linear speedup analysis of communication
  efficient momentum sgd for distributed non-convex optimization,'' in
  \emph{International Conference on Machine Learning (ICML)}.\hskip 1em plus
  0.5em minus 0.4em\relax PMLR, 2019, pp. 7184--7193.

\bibitem{xie2019local}
C.~Xie, O.~Koyejo, I.~Gupta, and H.~Lin, ``Local adaalter:
  Communication-efficient stochastic gradient descent with adaptive learning
  rates,'' \emph{OPT2020: 12th Annual Workshop on Optimization for Machine
  Learning}, 2020.

\bibitem{wang2021local}
J.~Wang, Z.~Xu, Z.~Garrett, Z.~Charles, L.~Liu, and G.~Joshi, ``Local
  adaptivity in federated learning: Convergence and consistency,''
  \emph{International Conference on Machine Learning (ICML)}, 2021.

\bibitem{reddi2020adaptive}
S.~J. Reddi, Z.~Charles, M.~Zaheer, Z.~Garrett, K.~Rush, J.~Kone{\v{c}}n{\`y},
  S.~Kumar, and H.~B. McMahan, ``Adaptive federated optimization,'' in
  \emph{International Conference on Learning Representations (ICLR)}.

\bibitem{wang2019slowmo}
J.~Wang, V.~Tantia, N.~Ballas, and M.~Rabbat, ``Slowmo: Improving
  communication-efficient distributed sgd with slow momentum,'' in
  \emph{International Conference on Learning Representations (ICLR)}.

\bibitem{chen2021cada}
T.~Chen, Z.~Guo, Y.~Sun, and W.~Yin, ``Cada: Communication-adaptive distributed
  adam,'' in \emph{International Conference on Artificial Intelligence and
  Statistics (AISTATS)}.\hskip 1em plus 0.5em minus 0.4em\relax PMLR, 2021, pp.
  613--621.

\bibitem{kairouz2021advances}
P.~Kairouz, H.~B. McMahan, B.~Avent, A.~Bellet, M.~Bennis, A.~N. Bhagoji,
  K.~Bonawitz, Z.~Charles, G.~Cormode, R.~Cummings \emph{et~al.}, ``Advances
  and open problems in federated learning,'' \emph{Foundations and
  Trends{\textregistered} in Machine Learning}, vol.~14, no. 1--2, pp. 1--210,
  2021.

\bibitem{article}
T.~D. Huynh, N.~Jennings, and N.~Shadbolt, ``Fire: An integrated trust and
  reputation model for open multi-agent systems,'' \emph{Journal of Autonomous
  Agents and Multi-Agent Systems}, vol.~13, pp. 18--22, 01 2004.

\bibitem{carrascosa2009service}
C.~Carrascosa, A.~Giret, V.~Julian, M.~Rebollo, E.~Argente, and V.~Botti,
  ``Service oriented mas: an open architecture,'' in \emph{Proceedings of The
  8th International Conference on Autonomous Agents and Multiagent
  Systems-Volume 2}, 2009, pp. 1291--1292.

\bibitem{8263752}
J.~M. Hendrickx and S.~Martin, ``Open multi-agent systems: Gossiping with
  random arrivals and departures,'' in \emph{IEEE Conference on Decision and
  Control (CDC)}, 2017, pp. 763--768.

\bibitem{hendrickx2020stability}
J.~M. Hendrickx and M.~G. Rabbat, ``Stability of decentralized gradient descent
  in open multi-agent systems,'' in \emph{IEEE Conference on Decision and
  Control (CDC)}, 2020, pp. 4885--4890.

\bibitem{hsieh2022multi}
Y.-G. Hsieh, F.~Iutzeler, J.~Malick, and P.~Mertikopoulos, ``Multi-agent online
  optimization with delays: Asynchronicity, adaptivity, and optimism,''
  \emph{Journal of Machine Learning Research (JMLR)}, vol.~23, no.~78, pp.
  1--49, 2022.

\bibitem{hsieh2021optimization}
------, ``Optimization in open networks via dual averaging,'' in \emph{IEEE
  Conference on Decision and Control (CDC)}, 2021, pp. 514--520.

\bibitem{lyapunov1992general}
A.~M. Lyapunov, ``The general problem of the stability of motion,''
  \emph{International journal of control}, vol.~55, no.~3, pp. 531--534, 1992.

\bibitem{kozin1969survey}
F.~Kozin, ``A survey of stability of stochastic systems,'' \emph{Automatica},
  vol.~5, no.~1, pp. 95--112, 1969.

\bibitem{hosoe2019equivalent}
Y.~Hosoe and T.~Hagiwara, ``Equivalent stability notions, lyapunov inequality,
  and its application in discrete-time linear systems with stochastic dynamics
  determined by an iid process,'' \emph{IEEE Transactions on Automatic
  Control}, vol.~64, no.~11, pp. 4764--4771, 2019.

\bibitem{hosoe2021second}
------, ``On second-moment stability of discrete-time linear systems with
  general stochastic dynamics,'' \emph{IEEE Transactions on Automatic Control},
  2021.

\bibitem{costa2013stochastic}
O.~Costa and D.~Z. Figueiredo, ``Stochastic stability of jump discrete-time
  linear systems with markov chain in a general borel space,'' \emph{IEEE
  Transactions on Automatic Control}, vol.~59, no.~1, pp. 223--227, 2013.

\bibitem{hardt2016train}
M.~Hardt, B.~Recht, and Y.~Singer, ``Train faster, generalize better: Stability
  of stochastic gradient descent,'' in \emph{International Conference on
  Machine Learning (ICML)}.\hskip 1em plus 0.5em minus 0.4em\relax PMLR, 2016,
  pp. 1225--1234.

\bibitem{bousquet2002stability}
O.~Bousquet and A.~Elisseeff, ``Stability and generalization,'' \emph{The
  Journal of Machine Learning Research (JMLR)}, vol.~2, pp. 499--526, 2002.

\bibitem{kingma2014adam}
D.~P. Kingma and J.~Ba, ``Adam: A method for stochastic optimization,''
  \emph{International Conference on Learning Representations (ICLR)}, 2015.

\bibitem{spiridonoff2020local}
A.~Spiridonoff, A.~Olshevsky, and I.~C. Paschalidis, ``Local sgd with a
  communication overhead depending only on the number of workers,'' 2020.

\bibitem{https://doi.org/10.48550/arxiv.2209.12307}
\BIBentryALTinterwordspacing
Y.~Sun, H.~Fernando, T.~Chen, and S.~Shahrampour, ``On the stability analysis
  of open federated learning systems,'' 2022. [Online]. Available:
  \url{https://arxiv.org/abs/2209.12307}
\BIBentrySTDinterwordspacing

\bibitem{deng2012mnist}
L.~Deng, ``The mnist database of handwritten digit images for machine learning
  research,'' \emph{IEEE Signal Processing Magazine}, vol.~29, no.~6, pp.
  141--142, 2012.

\end{thebibliography}
